\pdfoutput=1

\documentclass[11pt]{article}

\usepackage{acl}

\usepackage{times}
\usepackage{latexsym}

\usepackage[T1]{fontenc}

\usepackage[utf8]{inputenc}

\usepackage{microtype}

\usepackage{graphicx}
\usepackage{bm}
\usepackage{amsmath}
\usepackage{amsfonts}
\usepackage{booktabs}
\usepackage{multirow}
\usepackage{tikz}
\newcommand*\circled[1]{\tikz[baseline=(char.base)]{
            \node[shape=circle,draw,inner sep=0.9pt] (char) {#1};}}
\usepackage{subcaption}
\usepackage{makecell}

\title{Informative Language Representation Learning for Massively Multilingual Neural Machine Translation}

\author{Renren Jin \and Deyi Xiong \Thanks{~Corresponding author.} \\
        College of Intelligence and Computing, Tianjin University, Tianjin, China \\
        \texttt{\{rrjin, dyxiong\}@tju.edu.cn}}

\begin{document}
\maketitle

\begin{abstract}
In a multilingual neural machine translation model that fully shares parameters across all languages, an artificial language token is usually used to guide translation into the desired target language. However, recent studies show that prepending language tokens sometimes fails to navigate the multilingual neural machine translation models into right translation directions, especially on zero-shot translation. To mitigate this issue, we propose two methods, language embedding embodiment and language-aware multi-head attention, to learn informative language representations to channel translation into right directions. The former embodies language embeddings into different critical switching points along the information flow from the source to the target, aiming at amplifying translation direction guiding signals. The latter exploits a matrix, instead of a vector, to represent a language in the continuous space. The matrix is chunked into multiple heads so as to learn language representations in multiple subspaces. Experiment results on two datasets for massively multilingual neural machine translation demonstrate that language-aware multi-head attention benefits both supervised and zero-shot translation and significantly alleviates the off-target translation issue. Further linguistic typology prediction experiments show that matrix-based language representations learned by our methods are capable of capturing rich linguistic typology features.\footnote{The source code is publicly available at \url{https://github.com/cordercorder/nmt-multi}.}
\end{abstract}

\section{Introduction}

Multilingual neural machine translation (MNMT) \citep{DBLP:journals/tacl/JohnsonSLKWCTVW17,DBLP:conf/iwslt/HaNW16,DBLP:conf/naacl/AharoniJF19,DBLP:journals/corr/abs-1907-05019}, unlike bilingual machine translation with task-specific engineering  (language-specific features), uses a single learning system for multiple language pairs, which is jointly trained in a mult-task learning formalism \cite{DBLP:journals/jmlr/CollobertWBKKK11}. Parameter sharing across different languages in MNMT models enables transferring of intermediate representations and knowledge among languages, which makes it beneficial to machine translation of low-resource and even zero-resource languages \citep{DBLP:conf/emnlp/FiratSAYC16,DBLP:conf/naacl/GuHDL18,DBLP:conf/emnlp/NeubigH18,DBLP:conf/acl/GuWCL19,DBLP:conf/acl/ZhangWTS20}.

According to the degree of parameter sharing across languages, multilingual neural machine translation (MNMT) approaches can be categorized into two strands: full parameter sharing and partial parameter sharing \cite{DBLP:conf/wmt/SachanN18}. The former uses a unified model where both the encoder and decoder are shared for all languages. To guide the translation direction, a special token is usually prepended to the beginning of the source or target sentence to indicate the target language \citep{DBLP:journals/tacl/JohnsonSLKWCTVW17,DBLP:journals/jmlr/FanBSMEGBCWCGBL21}. Instead of sharing all model parameters for all translation directions, the latter uses language-specific components, e.g., separate encoders (in many-to-one translation), separate decoders (in one-to-many translation), or separate cross-attention networks \citep{DBLP:conf/naacl/ZophK16,DBLP:conf/naacl/FiratCB16,DBLP:conf/coling/BlackwoodBW18,DBLP:journals/coling/VazquezRCT20,DBLP:conf/eacl/KongRCTGL21}.

The key for full sharing models to distinguish target languages lies in the prepended tokens.\footnote{When there are multiple languages on the targe side, full sharing models cannot even converge during training if no target language information is provided by the prepended tokens \citep{DBLP:conf/acl/WangZZZXZ19}.} Prior studies find that the embeddings of the prepended tokens encode typological properties of languages \citep{DBLP:conf/eacl/TiedemannO17,DBLP:conf/emnlp/MalaviyaNL17,DBLP:journals/coling/BjervaOVTA19,DBLP:conf/emnlp/OncevayHB20}. Hence the token embeddings are also referred to as language representations or language embeddings, guiding translation into different target languages whose typological features varies a lot. While the prepened tokens play a significant role in language-specific knowledge learning, they are not usually working appropriately, making MNMT models generate translations in wrong languages, especially in zero-shot translation. Such off-target translation issue \citep{DBLP:conf/acl/ZhangWTS20,DBLP:conf/emnlp/YangEMTLH21} implies that prepending special tokens to sentence pairs in full-sharing models is not adequate to learn sufficient language-specific information to guide the MNMT models to translate into right directions.

Partial sharing may be suitable for learning language-specific properties. Nevertheless, it suffers from a rapid growth in the number of parameters with the increase in the number of languages. Additionally, which modules should be language-specific still remains to be further explored.

In this work, we attempt to learn more informative language representations (beyond language embeddings) to improve translation quality of the MNMT models with the full parameter sharing strategy. We argue that the prepended token cannot provide sufficient target language information to guide the translation direction, which is detrimental to both supervised and zero-shot translation. We conjecture that the direction control supervision provided by the prepended token is becoming weaker as the translation information flows to deeper layers. Therefore, instead of prepending special tokens to sentences, we propose a \underline{\textbf{L}}anguage \underline{\textbf{E}}mbedding \underline{\textbf{E}}mbodiment (LEE) strategy that embodies language embeddings at critical switching points (e.g., in-between self-attention and FFN, or self-attention and cross-attention) across layers in both the encoder and decoder along the information flow from the source to the target, so as to amplify the translation direction guiding signal, shown in Figure~\ref{transformer_lee}. Experiment results show a boosted translation performance compared with the standard full sharing MNMT model \citep{DBLP:journals/tacl/JohnsonSLKWCTVW17}.

Inspired by the performance improvement, we further propose \underline{\textbf{L}}anguage-\underline{\textbf{A}}ware Multi-Head \underline{\textbf{A}}ttention (LAA) to model typological features of languages. LAA estimates a matrix, instead of a vector, as the language representation for each language, which is able to encode more language information than a single fixed-length vector that is commonly adopted in previous studies. Motivated by multi-head attention \citep{DBLP:conf/nips/VaswaniSPUJGKP17}, we split the matrix along the column to learn information from different representation subspaces. In order to probe the typological properties encoded in language representations, we extract the language representations from the trained MNMT model and use them for linguistic typology prediction.

The main contributions of our work can be summarized as follows:
\begin{itemize}
    \item We empirically show that both the supervised and zero-shot translation of the standard full-sharing Transformer model are sensitive to the ways of indicating the desired target language.
    \item We propose the language embedding embodiment and language-aware multi-head attention to learn informative language representations for MNMT. We verify our proposal on two public datasets for massively MNMT in the many-to-many setting. Experimental results indicate that both the supervised and zero-shot translation can significantly benefit from LAA.
    \item We show that the language representations learned by LAA are generalizable and informative, which suggests that a proper language modeling strategy is crucial for MNMT.
\end{itemize}

\section{Related Work}
\paragraph{Multilingual Neural Machine Translation}
Pioneering studies on multilingual neural machine translation mostly favor to extend the standard bilingual model to MNMT by designing language specific components \citep{DBLP:conf/acl/DongWHYW15,DBLP:journals/corr/LuongLSVK15,DBLP:conf/naacl/ZophK16,DBLP:conf/naacl/FiratCB16}, where the number of parameters grows rapidly with the number of languages. Alternatively, \citet{DBLP:journals/tacl/JohnsonSLKWCTVW17,DBLP:conf/iwslt/HaNW16} propose to prepend an artificial target language token to source sentences without modifying the model architecture, which is parameter efficient and eases model design. It hence becomes the dominant approach to massively MNMT due to its simplicity and effectiveness \citep{DBLP:conf/naacl/AharoniJF19,DBLP:journals/corr/abs-1907-05019,DBLP:conf/wmt/FreitagF20,DBLP:conf/wmt/RiosMS20,DBLP:conf/acl/WuCWL21}. Despite that, it usually lags behind the bilingual counterpart on high-resource language pairs and suffers from off-target translation issue \citep{DBLP:conf/acl/ZhangWTS20} on zero-shot translation. To alleviate these issues, subsequent studies propose approaches such as increasing the model capacity by deepening the model \citep{DBLP:conf/acl/ZhangWTS20}, increasing the model cardinality \citep{DBLP:conf/acl/XuLGX20}, inserting mixture-of-experts (MoE) layers \citep{DBLP:journals/corr/abs-2101-03961}, designing lightweight language specific modules \citep{DBLP:conf/emnlp/WangZZXZ18,DBLP:conf/acl/WangZZZXZ19,DBLP:conf/coling/BlackwoodBW18,DBLP:conf/emnlp/BapnaF19,DBLP:conf/emnlp/PhilipBGB20,DBLP:conf/iclr/ZhangBSF21,DBLP:conf/emnlp/ZhuFZWL21}, reducing negative interference by clustering similar languages \citep{DBLP:conf/emnlp/TanCHXQL19}, resolving the gradient conflicts \citep{DBLP:conf/iclr/WangTF021}, dividing the model parameters into a shared part and language-specific part \citep{DBLP:conf/acl/LinWWL20,DBLP:conf/acl/Xie0G020,DBLP:journals/corr/abs-2104-07358,DBLP:conf/aaai/WangZ22}, bridging the representation gap by data argumentation \citep{DBLP:conf/emnlp/LinPWQFZL20} and contrastive learning \citep{DBLP:conf/acl/PanWWL20}, introducing language-agnostic regularization \citep{DBLP:journals/corr/abs-1903-07091,DBLP:conf/wmt/PhamNHW19} and explicit word alignment supervision \citep{DBLP:conf/emnlp/RaganatoVCT21}, to name a few. However, these studies neglect the capacity bottleneck in language representations as they all resort to a language embedding with the same dimension as word embeddings to encode the information of languages whose typological features diverse a lot. In contrast, LAA adopts the matrix as the language representation to alleviate the bottleneck. Moreover, the number of additional parameters brought by LAA is manageable thus it can adapt to massive MNMT settings.

\paragraph{Linguistic Typology Prediction}
Linguistic typology mainly studies the classification of languages based on their structural properties. Our work is closely related to inferring typological features with language representations from the trained MNMT models. Previous studies demonstrate that language representations in multilingual neural models can capture cross-lingual similarities between languages \citep{DBLP:conf/eacl/TiedemannO17,DBLP:conf/emnlp/MalaviyaNL17,DBLP:conf/naacl/BjervaA18,DBLP:journals/coling/BjervaOVTA19,DBLP:conf/acl/YuHS20}, which can potentially recover missing features in typological databases. \citet{DBLP:conf/emnlp/OncevayHB20} fuse language representations in trained MNMT models with features from typological databases to enhance knowledge transfer in MNMT models. They perform typological feature prediction to analyze typological knowledge in the learned language representations. Similarly, we conduct typological feature prediction to evaluate the quality of the language representations learned in LEE and LAA.

\section{Language Embedding Embodiment}

\begin{figure}[t]
    \centering
    \includegraphics[width=0.4\textwidth]{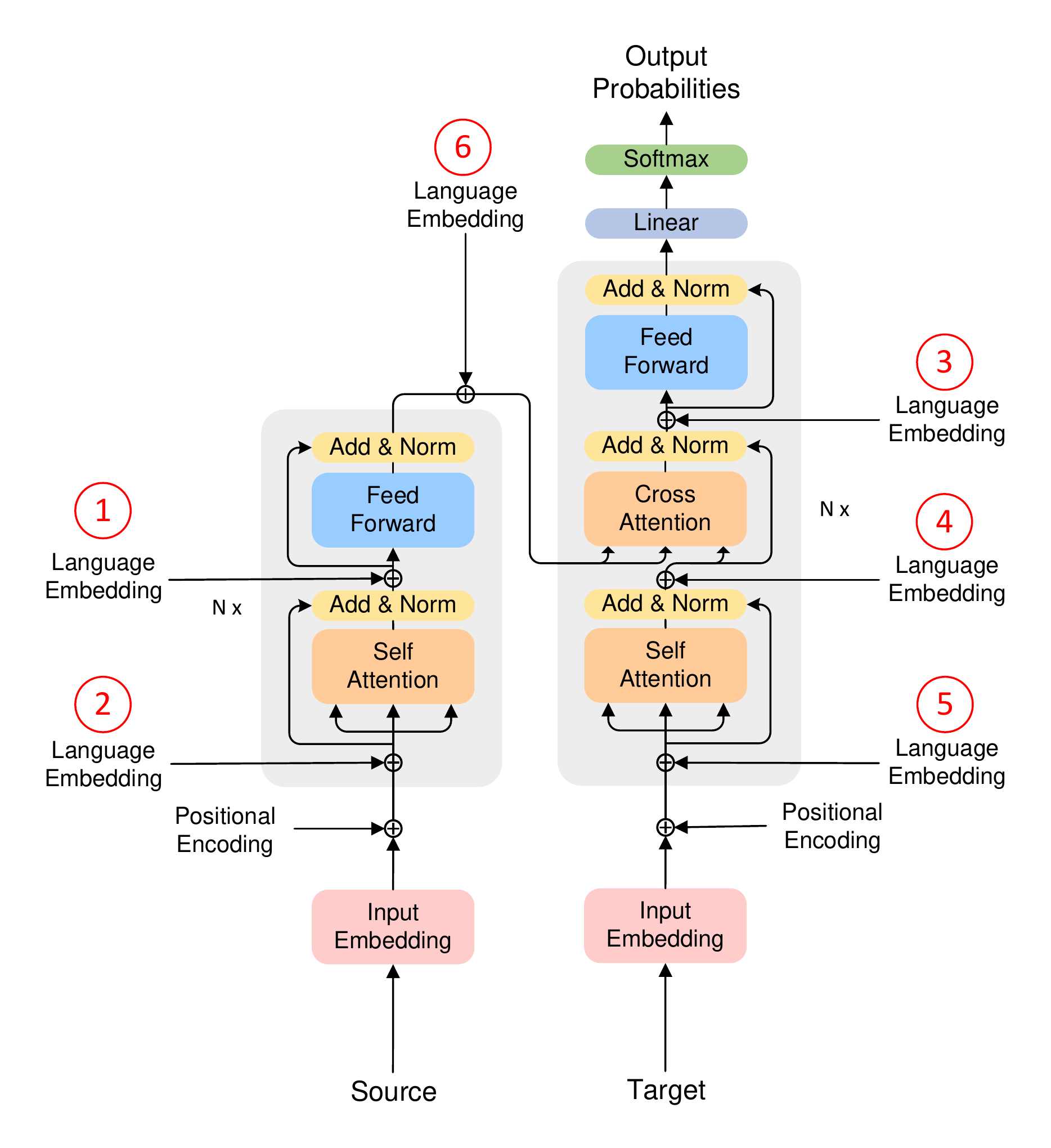}
    \caption{Illustration of the language embedding embodiment strategy. Six switching points in the standard Transformer are particularly marked to insert the language embedding.}\label{transformer_lee}
\end{figure}

The central idea for the language embedding embodiment strategy is shortening distance between the language embedding and target translation in order to enable language-embodied translation generation. In Transformer-based MNMT models, instead of only feeding the language embedding into the first layer \citep{DBLP:conf/nips/ConneauL19}, we embody the language embedding into many different layers along the translation information flow from the source to the target, as illustrated in Figure~\ref{transformer_lee}. Since the prepended artificial token has been already in vocabulary, feeding the language embedding into other layers will not result in extra parameters. As shown in Figure~\ref{transformer_lee}, six candidate positions in the standard Transformer can be used to embody the language embedding. Consider that the language embedding is to be embodied in position 5. Let $\bm{y}^{lang} = (\bm{y}_{1}^{lang}, \bm{y}_{2}^{lang}, \cdots, \bm{y}_{n}^{lang})$ denote the input of the target language $lang$ into position 5. $\bm{W}^{Q}$, $\bm{W}^{K}$ and $\bm{W}^{V}$ are parameter matrices of self-attention. $\bm{E}^{lang}$ is the language embedding of language $lang$. We compute the output $\bm{z}^{lang} = (\bm{z}_{1}^{lang}, \bm{z}_{2}^{lang}, \cdots, \bm{z}_{n}^{lang})$ after the language embedding is emodied as follows:
\begin{equation}\label{equ_1}
    e_{ij} = \frac{ ( ( \bm{y}_{i}^{lang} + \bm{E}^{lang} ) \bm{W}^{Q} ) ( ( \bm{y}_{j}^{lang} + \bm{E}^{lang} ) \bm{W}^{K} )^{\mathsf{T}}}{\sqrt{d}}
\end{equation}
\begin{equation}\label{equ_2}
    \bm{z}_{i}^{lang} = \sum_{j=1}^{n} \frac{\text{exp} ( e_{ij} )}{\sum_{k=1}^{n} \text{exp} ( e_{ik} )} ( \bm{y}_{j}^{lang} + \bm{E}^{lang} ) \bm{W}^{V}
\end{equation}
The language embedding embodiment for other positions can be done in a similar way: the language embedding is added to either word embeddings or hidden states fed into the corresponding position so as to make them language-specific. The embodied language embedding is helpful for the MNMT model to distinguish words/subwords that occur in different languages with different meanings.

\section{Language-Aware Multi-Head Attention}\label{Proposed Method}
\begin{figure}[t]
    \centering
    \includegraphics[width=0.4\textwidth]{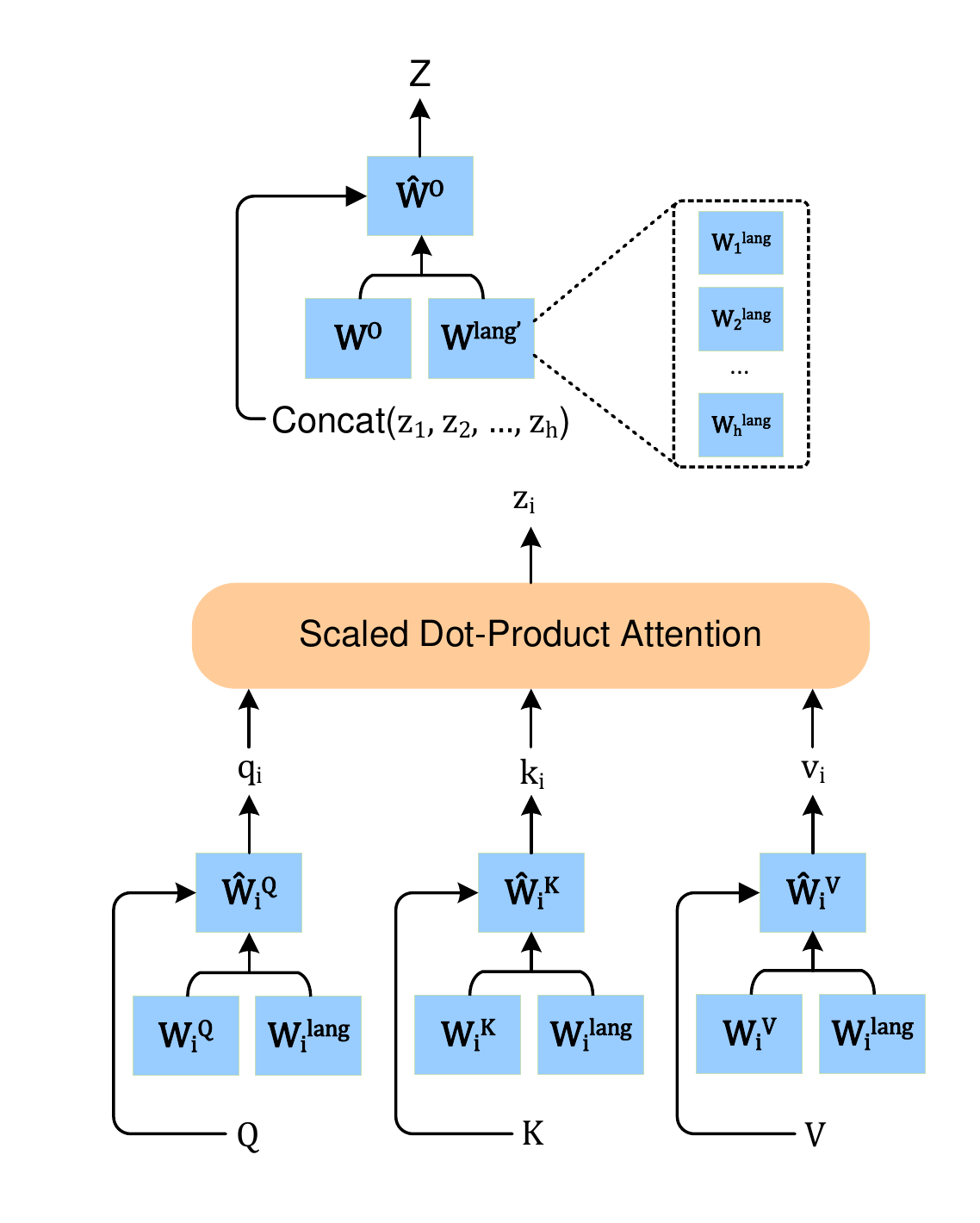}
    \caption{The architecture of the language-aware multi-head attention. $\bm{W}_{i}^{lang}$ is the trainable language-specific matrix of the $i$-th head.}\label{transformer_laa}
\end{figure}
LEE adopts a fixed-length language embedding to guide the MNMT model to generate translations of the target language. However, the fixed-length language vector may not be sufficient to capture all key linguistic features and diversities in languages, which are important for translation, due to its limited representational capacity. Experiment results of LEE on both zero-shot translation and linguistic typology prediction in Section~\ref{experiments} have also empirically verified the capacity bottleneck in the fixed-length language embedding. To alleviate the capacity bottleneck, we further propose the language aware multi-head attention (shown in Figure~\ref{transformer_laa}), representing a language with a matrix, rather than a vector. We incorporate the language-specific matrix into the attention modules (self-attention in the encoder/decoder or cross-attention) since they are the essential components in Transformer. Following the strategy of multi-head attention \citep{DBLP:conf/nips/VaswaniSPUJGKP17}, we split the matrix along its column to learn information from different representation subspaces of languages.

In the standard multi-head attention, three parameter matrices $\bm{W}_{i}^{Q}$, $\bm{W}_{i}^{K}$, $\bm{W}_{i}^{V}$ are used in the $i$-th attention head to project $\bm{Q}$ (queries), $\bm{K}$ (keys) and $\bm{V}$ (values). After projection, the scaled dot-product attention is applied on each head. The outputs from all heads are then concatenated and multiplied with $\bm{W}^{O}$. $\bm{W}_{i}^{Q} \in \mathbb{R}^{d_{model} \times d_{k}}$, $\bm{W}_{i}^{K} \in \mathbb{R}^{d_{model} \times d_{k}}$, $\bm{W}_{i}^{V} \in \mathbb{R}^{d_{model} \times d_{v}}$, $\bm{W}^{O} \in \mathbb{R}^{h d_{v} \times d_{model}}$ and $h$ is the number of heads. For simplicity, we set $d_{k} = d_{v} = d$ and $d_{model} = hd$. In practice, the parameter matrices of all heads from $\bm{Q}$, $\bm{K}$, $\bm{V}$ are concatenated into $\bm{W}^{Q}$, $\bm{W}^{K}$, $\bm{W}^{V}$ respectively and the projected results are split into $h$ heads, thus benefiting parallel computation and simplifying implementation.

In LAA, we use the matrix for language representation to relax the capacity constraint. Let matrix $\bm{W}^{lang}$ be the representation of language \textit{lang}, where $\bm{W}^{lang} \in \mathbb{R}^{d_{model} \times d_{model}}$. $\bm{W}^{lang}$ is split into $h$ heads along the column, which produces $\bm{W}_{i}^{lang} \in \mathbb{R}^{d_{model} \times d}$ for the $i$-th head. We add $\bm{W}_{i}^{lang}$ to $\bm{W}_{i}^{Q}$, $\bm{W}_{i}^{K}$, $\bm{W}_{i}^{V}$ to inject the target language information into MNMT models, which can be formulated as follows:
\begin{equation}\label{equ_3}
    \begin{split}
        \bm{q}_{i} = \bm{Q} (\bm{W}_{i}^{Q} + \bm{W}_{i}^{lang})\\ 
        \bm{k}_{i} = \bm{K} (\bm{W}_{i}^{K} + \bm{W}_{i}^{lang})\\
        \bm{v}_{i} = \bm{V} (\bm{W}_{i}^{V} + \bm{W}_{i}^{lang})
    \end{split}
\end{equation}
Let $\bm{z}_{i}$ be the scaled dot-product attention output of the $i$-th head. $\bm{z}_{i}$ is computed as:
\begin{equation}\label{equ_4}
    \bm{z}_{i} = \text{softmax} ( \frac{\bm{q}_{i}\bm{k}_{i}^{\mathsf{T}}}{\sqrt{d}} ) \bm{v}_{i}
\end{equation}
We split $\bm{W}^{O}$ into $h$ heads along the row and pack the outputs from different heads together as follows:
\begin{equation}\label{equ_5}
    \bm{Z} = \sum_{i=1}^{h} \bm{z}_{i} (\bm{W}_{i}^{O} + ({\bm{W}_{i}^{lang}})^{\mathsf{T}})
\end{equation}
where $\bm{W}_{i}^{O} \in \mathbb{R}^{d \times d_{model}}$.
\begin{table*}[t]
\centering
    \resizebox{\textwidth}{!}{
        \begin{tabular}{c|c|c|c|c|c}
        \toprule
            {Dataset} & {Type} & {\#Languages} & {\#Supervised directions} & {\#Zero-shot directions} & {\#Training sentences} \\
            \midrule
            {TED-59} & English-centric & 59 & 116 & 3306 & 5M \\
        \midrule
            {OPUS-100} & English-centric & 100 & 198 & 30 & 55M\\
        \bottomrule
        \end{tabular}
    }
\caption{Statistics of the two datasets.}\label{dataset_summary}
\end{table*}
As Eq.~(\ref{equ_3}) and (\ref{equ_5}) show, the language representation matrix can be incorporated into the self-attention layer of the encoder/decoder or the cross-attention layer. It is optimized with other parameters to capture the typological properties of a given language. The MNMT models with the matrix-based $\bm{W}^{lang}$ learned in this way, have sufficient representation capacity to model both language-specific features (so as to alleviate the negative interference \citep{DBLP:conf/iclr/WangTF021} between dissimilar languages) and universal features shared by all languages. Additionally, LAA is able to scale to massively MNMT as there is only one additional matrix for each language and it introduces a very small extra computational cost in training and inference. We show several tips to implement LAA efficiently in Appendix~\ref{efficient_implementation_of_laa}.

\section{Experiments}\label{experiments}
We conducted two types of experiments, many-to-many translation and linguistic typology prediction, aiming at: (1) systematically evaluating the traditional language token prepending method and (2) examing the effectiveness of the proposed language embedding embodiment and language-aware multi-head attention.

\subsection{Experiment Settings for Many-to-Many Translation}\label{experimental_setup}
\begin{table*}[t]
\centering
    \resizebox{\textwidth}{!}{
        \begin{tabular}{c|c|l|c|cc|cc|cc|ccc}
        \toprule
            \multicolumn{1}{c}{\multirow{2}{*}{ID}} & \multicolumn{1}{c}{\multirow{2}{*}{Dataset}} & \multicolumn{1}{c}{\multirow{2}{*}{Model}} & \multicolumn{1}{c}{\multirow{2}{*}{\#Param}} & \multicolumn{2}{c|}{En $\rightarrow$ XX} &  \multicolumn{2}{c|}{XX $\rightarrow$ En} & \multicolumn{2}{c|}{All} & \multicolumn{3}{c}{Zero-shot} \\
        \cmidrule{5-13}
            \multicolumn{1}{c}{} & \multicolumn{1}{c}{} & \multicolumn{1}{c}{} & \multicolumn{1}{c}{} & BLEU & WR & BLEU & WR & BLEU & WR & BLEU & LangAcc & WR \\
        \midrule
            \circled{1} & TED-59 & Token$_{\mathrm{tgt}}$ & 77M & 20.74 & \textit{ref} & 24.08 & \textit{ref} & 22.41 & \textit{ref} & 2.42 & 37.62 & \textit{ref} \\
            \circled{2} & TED-59 & Token$_{\mathrm{src}}$ & 77M & 21.24 & \textbf{91.38} & 23.77 & 15.52 & 22.50 & 53.45 & \textbf{10.50} & 71.82 & \textbf{98.91} \\
            \circled{3} & TED-59 & LEE$_{2}$ & 77M & \textbf{21.26} & 89.66 & 23.42 & 5.17 & 22.34 & 47.41 & 8.87 & 73.97 & 97.70 \\
            \circled{4} & TED-59 & LEE$_{1,2}$ & 77M & 20.92 & 65.52 & 23.75 & 32.76 & 22.33 & 49.14 & 8.99 & 73.97 & 97.58 \\
            \circled{5} & TED-59 & LEE$_{5}$ & 77M & 20.70 & 48.28 & 24.32 & 77.59 & 22.51 & 62.93 & 4.10 & 58.63 & 83.42 \\
            \circled{6} & TED-59 & LEE$_{4,5}$ & 77M & 20.83 & 65.52 & 24.49 & 84.48 & \textbf{22.66} & \textbf{75.00} & 3.94 & 59.60 & 75.05 \\
            \circled{7} & TED-59 & LEE$_{3,4,5}$ & 77M & 20.65 & 39.66 & \textbf{24.56} & \textbf{93.10} & 22.61 & 66.38 & 4.98 & 61.76 & 87.69 \\
            \circled{8} & TED-59 & LEE$_{4,5,6}$ & 77M & 20.71 & 51.72 & 24.39 & 89.66 & 22.55 & 70.69 & 6.17 & 64.83 & 94.65 \\
            \circled{9} & TED-59 & LEE$_{2,5}$ & 77M & \textbf{21.26} & 86.21 & 23.79 & 27.59 & 22.53 & 56.90 & 9.82 & \textbf{74.44} & 98.55 \\
        \midrule
            \circled{\small 10} & OPUS-100 & Token$_{\mathrm{tgt}}$ & 77M & 23.46 & \textit{ref} & 29.49 & \textit{ref} & 26.47 & \textit{ref} & 5.81 & 55.92 & \textit{ref} \\
            \circled{\small 11} & OPUS-100 & Token$_{\mathrm{src}}$ & 77M & \textbf{24.04} & \textbf{81.91} & 28.74 & 8.51 & 26.39 & 45.21 & 4.06 & 34.53 & 23.33 \\
            \circled{\small 12} & OPUS-100 & LEE$_{4,5}$ & 77M & 23.38 & 50.00 & \textbf{29.64} & 47.87 & \textbf{26.51} & 48.94 & \textbf{7.37} & \textbf{65.84} & \textbf{56.67} \\
        \bottomrule
        \end{tabular}
    }
\caption{Experimental results of LEE on the two datasets. En $\rightarrow$ XX and XX $\rightarrow$ En: translation directions from and to English respectively. All: all supervised translation directions. Zero-shot: zero-shot translation directions. \#Param: the number of trainable parameters in the model. WR (Win Rate): the proportion of translation directions that outperform the \textit{ref} system in terms of BLEU. LangAcc: the proportion of translations in the correct target language among all translations. Token$_{\mathrm{tgt}}$: prepending a target language tag to the target side. Token$_{\mathrm{src}}$: prepending a target language tag to the source side. The subscript of LEE denotes the positions to embody the language embedding as shown in Figure~\ref{transformer_lee}. It's worth noting that for supervised translation directions, the test sets in OPUS-100 only cover 94 language pairs and BLEU for "All" is hence averaged on 188 translation directions.}\label{pre_mnmt_results}
\end{table*}
\begin{table*}[t]
\centering
    \resizebox{\textwidth}{!}{
        \begin{tabular}{c|c|l|c|cc|cc|cc|ccc}
        \toprule
            \multicolumn{1}{c}{\multirow{2}{*}{ID}} & \multicolumn{1}{c}{\multirow{2}{*}{Dataset}} & \multicolumn{1}{c}{\multirow{2}{*}{Model}} & \multicolumn{1}{c}{\multirow{2}{*}{\#Param}} & \multicolumn{2}{c|}{En $\rightarrow$ XX} &  \multicolumn{2}{c|}{XX $\rightarrow$ En} & \multicolumn{2}{c|}{All} & \multicolumn{3}{c}{Zero-shot} \\
        \cmidrule{5-13}
            \multicolumn{1}{c}{} & \multicolumn{1}{c}{} & \multicolumn{1}{c}{} & \multicolumn{1}{c}{} & BLEU & WR & BLEU & WR & BLEU & WR & BLEU & LangAcc & WR \\
        \midrule
            \circled{1} & TED-59 & Token$_{\mathrm{tgt}}$ & 77M & 20.74 & \textit{ref} & 24.08 & \textit{ref} & 22.41 & \textit{ref} & 2.42 & 37.62 & \textit{ref} \\
            \circled{2} & TED-59 & LAA$_{\mathrm{enc.self}}$ & 92M & 20.11 & 17.24 & 21.13 & 1.72 & 20.62 & 9.48 & 6.30 & 71.92 & 92.77 \\
            \circled{3} & TED-59 & LAA$_{\mathrm{dec.self}}$ & 92M & 21.29 & 84.48 & 25.14 & 96.55 & 23.21 & 90.52 & 8.94 & 74.68 & 98.58 \\
            \circled{4} & TED-59 & LAA$_{\mathrm{enc.self, dec.self}}$ & 92M & 21.04 & 68.97 & 23.10 & 1.72 & 22.07 & 35.34 & 8.89 & 74.03 & 97.58 \\
            \circled{5} & TED-59 & LAA$_{\mathrm{dec.self,cros}}$ & 92M & \textbf{21.47} & 89.66 & 24.79 & 94.83 & 23.13 & 92.24 & \textbf{12.69} & 74.57 & \textbf{99.15} \\
            \circled{6} & TED-59 & LAA$_{\mathrm{dec.self}}$ + Token$_{\mathrm{tgt}}$ & 92M & 21.19 & 79.31 & 25.11 & 96.55 & 23.15 & 87.93 & 9.07 & 74.65 & 98.40 \\
            \circled{7} & TED-59 & LAA$_{\mathrm{dec.self}}$ + LEE$_{4,5}$ & 92M & 21.28 & 87.93 & \textbf{25.29} & \textbf{100.00} & \textbf{23.28} & \textbf{93.97} & 9.48 & 74.69 & 99.09 \\
            \circled{8} & TED-59 & Adapter$_{\mathrm{enc}}$ & 92M & 21.04 & 68.97 & 22.91 & 0.00 & 21.97 & 34.48 & 8.20 & 74.44 & 96.79 \\
            \circled{9} & TED-59 & Adapter$_{\mathrm{dec}}$ & 92M & 21.43 & \textbf{91.38} & 24.14 & 62.07 & 22.79 & 76.72 & 10.61 & \textbf{75.36} & 98.97 \\
            \circled{\small 10} & TED-59 & Adapter$_{\mathrm{enc,dec}}$ & 92M & 21.10 & 79.31 & 22.95 & 1.72 & 22.03 & 40.52 & 9.17 & 75.10 & 98.09 \\
            \circled{\small 11} & TED-59 & LALN + LALT & 94M & 21.37 & 75.86 & 22.85 & 1.72 & 22.11 & 38.79 & 7.69 & 73.18 & 95.58 \\
        \midrule
            \circled{\small 12} & OPUS-100 & Token$_{\mathrm{tgt}}$ & 77M & 23.46 & \textit{ref} & 29.49 & \textit{ref} & 26.47 & \textit{ref} & 5.81 & 55.92 & \textit{ref} \\
            \circled{\small 13} & OPUS-100 & LAA$_{\mathrm{dec.self}}$ & 103M & \textbf{24.42} & \textbf{90.43} & 29.78 & \textbf{79.79} & 27.10 & \textbf{85.11} & 11.23 & 81.53 & \textbf{100.00} \\
            \circled{\small 14} & OPUS-100 & LAA$_{\mathrm{dec.self,cros}}$ & 103M & \textbf{24.42} & 89.36 & 29.58 & 68.09 & 27.00 & 78.72 & \textbf{12.13} & \textbf{87.65} & \textbf{100.00} \\
            \circled{\small 15} & OPUS-100 & LAA$_{\mathrm{dec.self}}$ + LEE$_{4,5}$ & 103M & 24.41 & 88.30 & \textbf{29.83} & 76.60 & \textbf{27.12} & 82.45 & 11.31 & 81.70 & \textbf{100.00} \\
        \bottomrule
        \end{tabular}
    }
\caption{Experiment results of LAA on the two datasets. The results of \circled{1} and \circled{\small 12} are from Table~\ref{pre_mnmt_results}. The subscript of LAA denotes the layers where the language representation is incorporated. The subscript of the adapter denotes the modules where the monolingual adapter is incorporated. enc/dec denote the encoder/decoder. self/cros denote self-attention/cross-attention layer. Considering that both LAA$_{\mathrm{dec.self}}$ and LAA$_{\mathrm{dec.self}}$ + LEE$_{4,5}$ achieve superior average BLEU scores on supervised translation directions and LAA$_{\mathrm{dec.self,cros}}$ obtains the best BLEU on zero-shot translation, we verify them again on the OPUS-100 dataset.}\label{mnmt_results}
\end{table*}
\begin{table*}[t]
\centering
    \resizebox{\textwidth}{!}{
        \begin{tabular}{c|c|l|c|cc|cc|cc|ccc}
        \toprule
            \multicolumn{1}{c}{\multirow{2}{*}{ID}} & \multicolumn{1}{c}{\multirow{2}{*}{Dataset}} & \multicolumn{1}{c}{\multirow{2}{*}{Model}} & \multicolumn{1}{c}{\multirow{2}{*}{\#Param}} & \multicolumn{2}{c|}{En $\rightarrow$ XX} &  \multicolumn{2}{c|}{XX $\rightarrow$ En} & \multicolumn{2}{c|}{All} & \multicolumn{3}{c}{Zero-shot} \\
        \cmidrule{5-13}
            \multicolumn{1}{c}{} & \multicolumn{1}{c}{} & \multicolumn{1}{c}{} & \multicolumn{1}{c}{} & BLEU & WR & BLEU & WR & BLEU & WR & BLEU & LangAcc & WR \\
        \midrule
            \circled{1} & TED-59 & Token$_{\mathrm{tgt}}^{-}$ & 77M & 19.54 & \textit{ref} & 24.23 & \textit{ref} & 21.89 & \textit{ref} & 2.84 & 40.94 & \textit{ref} \\
            \circled{2} & TED-59 & Token$_{\mathrm{src}}^{-}$ & 77M & 20.25 & \textbf{96.55} & 23.65 & 8.62 & 21.95 & 52.59 & 9.65 & 65.45 & 96.77 \\
            \circled{3} & TED-59 & LEE$_{4,5}^{-}$ & 77M & 19.56 & 51.72 & 24.42 & 86.21 & 21.99 & 68.97 & 6.43 & 66.25 & 87.57 \\
            \circled{4} & TED-59 & LAA$_{\mathrm{dec.self}}^{-}$ & 92M & 20.64 & \textbf{96.55} & \textbf{25.16} & \textbf{98.28} & \textbf{22.90} & \textbf{97.41} & 9.93 & 75.04 & 97.67 \\
            \circled{5} & TED-59 & LAA$_{\mathrm{dec.self}}^{-}$ + LEE$_{4,5}^{-}$ & 92M & \textbf{20.67} & 94.83 & 25.05 & 93.10 & 22.86 & 93.97 & \textbf{10.02} & \textbf{75.26} & \textbf{97.82} \\
        \midrule
            \circled{6} & OPUS-100 & \citet{DBLP:conf/acl/ZhangWTS20} & 254M & 23.96 & - & 31.36 & - & 27.66 & - & 5.24 & 47.91 & - \\
            \circled{7} & OPUS-100 & \citet{DBLP:conf/acl/ZhangWTS20} & 254M & 23.36 & - & 30.98 & - & 27.17 & - & 14.08 & \textbf{87.68} & - \\
            \circled{8} & OPUS-100 & \citet{DBLP:conf/acl/XuLGX20} & - & \textbf{24.17} & - & \textbf{32.19} & - & \textbf{28.18} & - & \textbf{14.71} & - & - \\
        \midrule
            \circled{9} & OPUS-100 & Token$_{\mathrm{tgt}}^{-}$ & 77M & 21.82 & \textit{ref} & 28.45 & \textit{ref} & 25.14 & \textit{ref} & 6.63 & 58.76 & \textit{ref} \\
            \circled{\small 10} & OPUS-100 & Token$_{\mathrm{src}}^{-}$ & 77M & 22.15 & 74.47 & 27.68 & 10.64 & 24.91 & 42.55 & 4.91 & 37.80 & 30.00 \\
            \circled{\small 11} & OPUS-100 & LEE$_{4,5}^{-}$ & 77M & 21.49 & 45.74 & 28.48 & 43.62 & 24.98 & 44.68 & 10.08 & 79.90 & 96.67 \\
            \circled{\small 12} & OPUS-100 & LAA$_{\mathrm{dec.self}}^{-}$ & 103M & 23.57 & 91.49 & 28.71 & 70.21 & 26.14 & 80.85 & 11.93 & 80.39 & \textbf{100.00} \\
            \circled{\small 13} & OPUS-100 & LAA$_{\mathrm{dec.self}}^{-}$ + LEE$_{4,5}^{-}$ & 103M & 23.69 & 91.49 & 28.88 & 78.72 & 26.29 & 85.11 & 12.77 & \textbf{85.00} & \textbf{100.00} \\
            \circled{\small 14} & OPUS-100 & \circled{\small 13} + 24 layers & 236M & \textbf{26.82} & \textbf{98.94} & \textbf{32.31} & \textbf{100.00} & \textbf{29.56} & \textbf{99.47} & \textbf{15.08} & 84.59 & \textbf{100.00} \\
        \bottomrule
        \end{tabular}
    }
\caption{Experimental results without oversampling on the two datasets. The superscript "-" denotes that oversampling is removed.}\label{mnmt_results_no_over_sample}
\end{table*}
\paragraph{Dataset} We used two publicly available MNMT datasets TED-59 \citep{DBLP:conf/naacl/QiSFPN18} and OPUS-100 \citep{DBLP:conf/acl/ZhangWTS20}. Table~\ref{dataset_summary} shows the statistics of the two datasets. Due to the multi-way parallel nature of the TED-59 dataset,\footnote{Each English sentence may have multiple corresponding translations in different languages.} we were able to construct 3306 ($58 \times 57$) test sets for zero-shot translation directions by pairing sentences of any two languages via aligned English sentences in the original test sets. To balance training data for various language pairs, we employed the temperature-based sampling strategy \citep{DBLP:journals/corr/abs-1907-05019} with $T = 5$. More details are in Appendix~\ref{dataset_for_many_to_many_translation}.
\paragraph{Model \& Training}
We built our MNMT models on the Transformer base. More detailed settings are shown in Appendix~\ref{mnmt_model_training_appendix}.
\paragraph{Evaluation}
We adopted case-sensitive de-tokenized sacreBLEU\footnote{For TED-59, the signature is: BLEU+case.mixed+numre-\\fs.1+smooth.exp+tok.\{13a,zh,ja-mecab-0.996\}+version.1.5-\\.1, tok.zh and tok.ja-mecab-0.996 are only for Chinese and Japanese respectively. For OPUS-100, We adopted the same signature as previous work to be comparable with them: BLEU+case.mixed+numrefs.1+smooth.exp+tok.13a+versio-\\n.1.4.1} \citep{DBLP:conf/wmt/Post18} as the evaluation metric. BLEU scores were averaged over test sets. Following \citet{DBLP:conf/acl/ZhangWTS20}, we used LangAcc as a complementary evaluation metric for zero-shot translation, which calculates the proportion of the translations in the desired target language. More details about the evaluation settings are in Appendix~\ref{mnmt_model_evaluation_appendix}.
\subsection{Experiment Settings for Linguistic Typology Prediction}
\paragraph{Dataset}
We employed typological features from URIEL typological database \citep{DBLP:conf/eacl/LevinLMLKT17}\footnote{\url{https://www.cs.cmu.edu/~dmortens/projects/07_project}} for experiments. We mainly used syntax, phonology and phonetic inventory typological features in our work. More details are in Appendix~\ref{dataset_for_linguistic_typology_prediction}.

\paragraph{Prediction Methods}
We adopted $k$-nearest neighbors approach (k-NN) for linguistic typology prediction. More details about the prediction methods are in Appendix~\ref{prediction_method_for_typology_features}.

\subsection{Many-to-Many Translation Results}
We conducted a series of experiments to evaluate both LEE and LAA in many-to-many translation.
\subsubsection{Language Embedding Embodiment}\label{experimental_results_mnmt_lee}
We mainly carried out experiments on the TED-59 dataset to examine the effectiveness of LEE. Table~\ref{pre_mnmt_results} shows the experimental results. We further evaluated LEE$_{4,5}$ (the best system in terms of average BLEU on all supervised translation directions on the TED-59 dataset) on the OPUS-100 dataset. As shown in Table~\ref{pre_mnmt_results}, poor LangAccs on zero-shot translation directions suggest that the majority of models suffer from the severe off-target translation issue \citep{DBLP:conf/acl/ZhangWTS20}. Besides, LEE$_{4,5}$ can slightly boost the average BLEU on supervised translation directions on both datasets. Findings from Table~\ref{pre_mnmt_results} can be summarized as follows:
\begin{itemize}
    \item The ways to indicate the desired target language to MNMT models can significantly affect the translation performance, especially on zero-shot translation directions.
    \item The performance gap between Token$_{\mathrm{src}}$ and Token$_{\mathrm{tgt}}$ on zero-shot translation directions varies substantially across different datasets. We perform analysis to find the main reasons behind the performance discrepancy and the details are shown in Appendix~\ref{detailed_many_to_many_translation_results_for_lee}.
    \item Prepending a target language token to the source side (Token$_{\mathrm{src}}$) and to the target side (Token$_{\mathrm{tgt}}$) benefit En $\rightarrow$ XX and XX $\rightarrow$ En translation directions respectively.
    \item Embodying the target language embedding in the decoder is preferable for supervised translation directions.
\end{itemize}
Please refer to Appendix~\ref{detailed_many_to_many_translation_results_for_lee} for an in-depth analysis that supports these findings.

\subsubsection{Language-Aware Multi-Head Attention}
\paragraph{Baslines}
Since LAA enlarges the language representation with a language-specific matrix, introducing additional parameters than LEE, we compared it against two state-of-the-art MNMT baselines. The first is the monolingual adapter \citep{DBLP:conf/emnlp/PhilipBGB20}. We omit the bottleneck dimension and activation function in the original adapter and adopt a matrix as the adapter. The adapter is the same size as the language representation in LAA. We share the monolingual adapter across all layers to keep the number of parameters identical to LAA and jointly train it with the other components of the MNMT model. In addition to the monolingual adapter, we also compared against the combination of LALN (Language-Aware Layer Normalization) and LALT (Language-Aware Linear Transformation) proposed by \citet{DBLP:conf/acl/ZhangWTS20}.

Table~\ref{mnmt_results} shows the detailed results. We observe that incorporating LAA into the decoder benefits both supervised and zero-shot translation (\circled{1} vs. \circled{3}\circled{5})(\circled{2} vs. \circled{4})(\circled{\small 12} vs. \circled{\small 13}\circled{\small 14}), while incorporating LAA into the self-attention layer of the encoder deteriorates translation performance (\circled{1} vs. \circled{2})(\circled{3} vs. \circled{4}), which is consistent with the finding in LEE (i.e., the incorporation into the the decoder is better than the encoder). Similarly, incorporating the monolingual adapter into the decoder is more effective than into the encoder (\circled{8} vs. \circled{9}). \textbf{Moreover, LAA outperforms all strong baselines under the condition of using equal number of parameters} (\circled{8}\circled{9}\circled{\small 10}\circled{\small 11} vs. \circled{3}\circled{5}), which demonstrates that LAA can capture rich language information to improve translation performance of massively MNMT model. To examine whether LAA and LEE are complementary to each other, we combine LAA$_{\mathrm{dec.self}}$ and LEE$_{4,5}$ and the results show a slightly boosted translation performance (\circled{3} vs. \circled{7}) (\circled{\small 13} vs. \circled{\small 15}). However, adopting LAA while prepending the language tag to the target side leads to a slightly performance drop on supervised translation directions (\circled{3} vs. \circled{6}).

The experiments in Table~\ref{pre_mnmt_results} and \ref{mnmt_results} are conducted under a temperature-based sampling setting of $T=5$, which oversamples training data of low-resource language pairs. Oversampling usually leads to improvements in average BLEU on supervised translation directions (Table~\ref{mnmt_results} vs. Table~\ref{mnmt_results_no_over_sample}). To eliminate such effect and compare with previous work \citep{DBLP:conf/acl/ZhangWTS20, DBLP:conf/acl/XuLGX20},\footnote{As the two studies are carried out on the OPUS-100 dataset without oversampling, we remove oversampling to make a fair comparison with them.} we carried out experiments without oversampling. Table~\ref{mnmt_results_no_over_sample} summarizes our experimental results without oversampling. The results suggest that LAA yields larger BLEU improvements on supervised translation directions when training on the raw data distribution. e.g., (\circled{\small 13} - \circled{\small 12} = +0.63 BLEU, Table~\ref{mnmt_results}) vs. (\circled{\small 12} - \circled{9} = +1 BLEU, Table~\ref{mnmt_results_no_over_sample}). Additionally, the performance gain on zero-shot translation is also retained mostly (\circled{\small 13} - \circled{\small 12} = +5.42 BLEU, Table~\ref{mnmt_results}) vs. (\circled{\small 12} - \circled{9} = +5.3 BLEU, Table~\ref{mnmt_results_no_over_sample}). We enlarge the MNMT model by increasing the number of layers in both the encoder and decoder to 24 so that the number of parameters in our models is approximately the same as the largest model used by \citet{DBLP:conf/acl/ZhangWTS20} and  \citet{DBLP:conf/acl/XuLGX20}.\footnote{Although \citet{DBLP:conf/acl/XuLGX20} don't report the exact number of parameters, the number of parameters in their model is approximately equal to that of \citet{DBLP:conf/acl/ZhangWTS20} as stated in paper.} As the model deepens, we adjusted the dropout rate to 0.2. Our methods outperform previous results \citep{DBLP:conf/acl/ZhangWTS20,DBLP:conf/acl/XuLGX20} on both supervised and zero-shot translation directions (\circled{6}\circled{7}\circled{8} vs. \circled{\small 14}).\footnote{\citet{DBLP:conf/emnlp/YangEMTLH21} and \citet{DBLP:journals/corr/abs-2203-00555} report better results than us. However, \citet{DBLP:conf/emnlp/YangEMTLH21} oversampled the training data for low-resource language pairs and removed five language pairs without test data. We removed oversampling and used all training data of the OPUS-100 dataset, hence the model in our work needs to schedule its capacity over the extra ten translation directions. \citet{DBLP:journals/corr/abs-2203-00555} conducted experiments with the MNMT model of up to 3.8B parameters, which is several times larger than ours. To the best of our knowledge, there are no other results superior to ours except theirs.}

\subsection{Linguistic Typology Prediction Results}
Similarly, we conducted experiments on the linguistic typology prediction for both LEE and LAA which are trained on the two datasets.
\begin{figure*}[t]
    \centering
        \begin{subfigure}{0.325\textwidth}
            \centering
            \includegraphics[width=\linewidth]{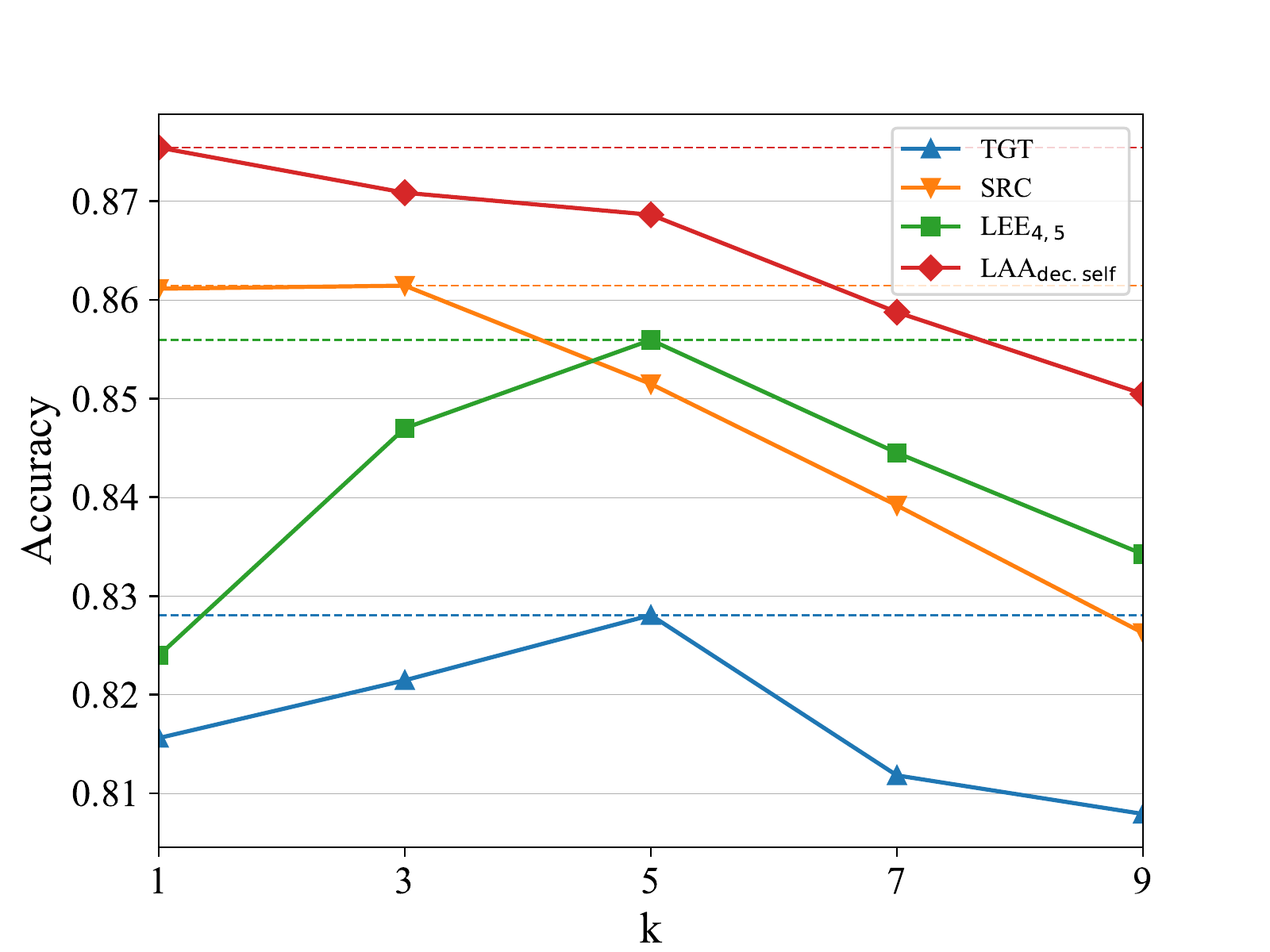}  
            \caption{Syntax}
        \end{subfigure}
        \begin{subfigure}{0.325\textwidth}
            \centering
            \includegraphics[width=\linewidth]{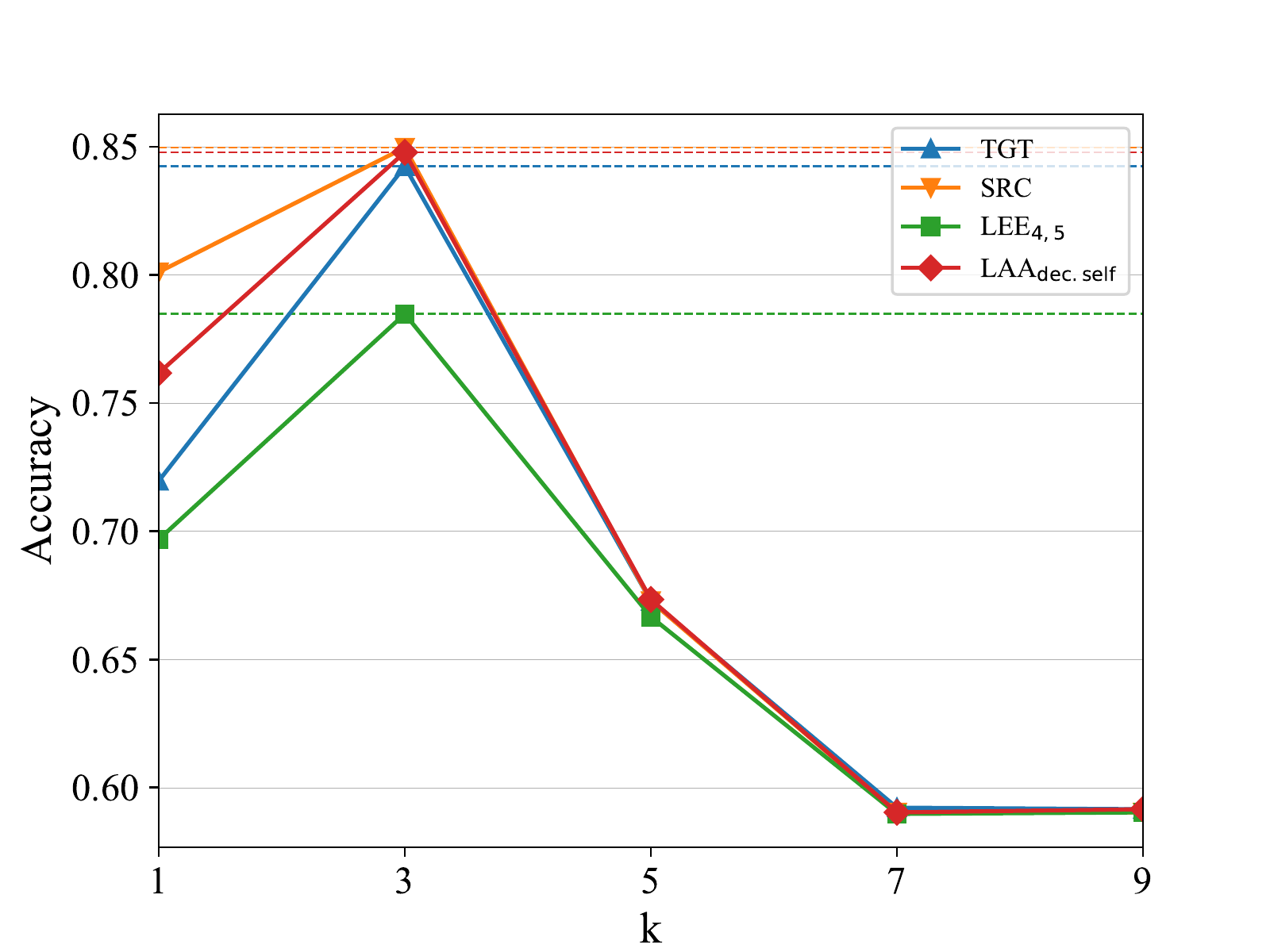}  
            \caption{Phonology}
        \end{subfigure}
        \begin{subfigure}{0.325\textwidth}
            \centering
            \includegraphics[width=\linewidth]{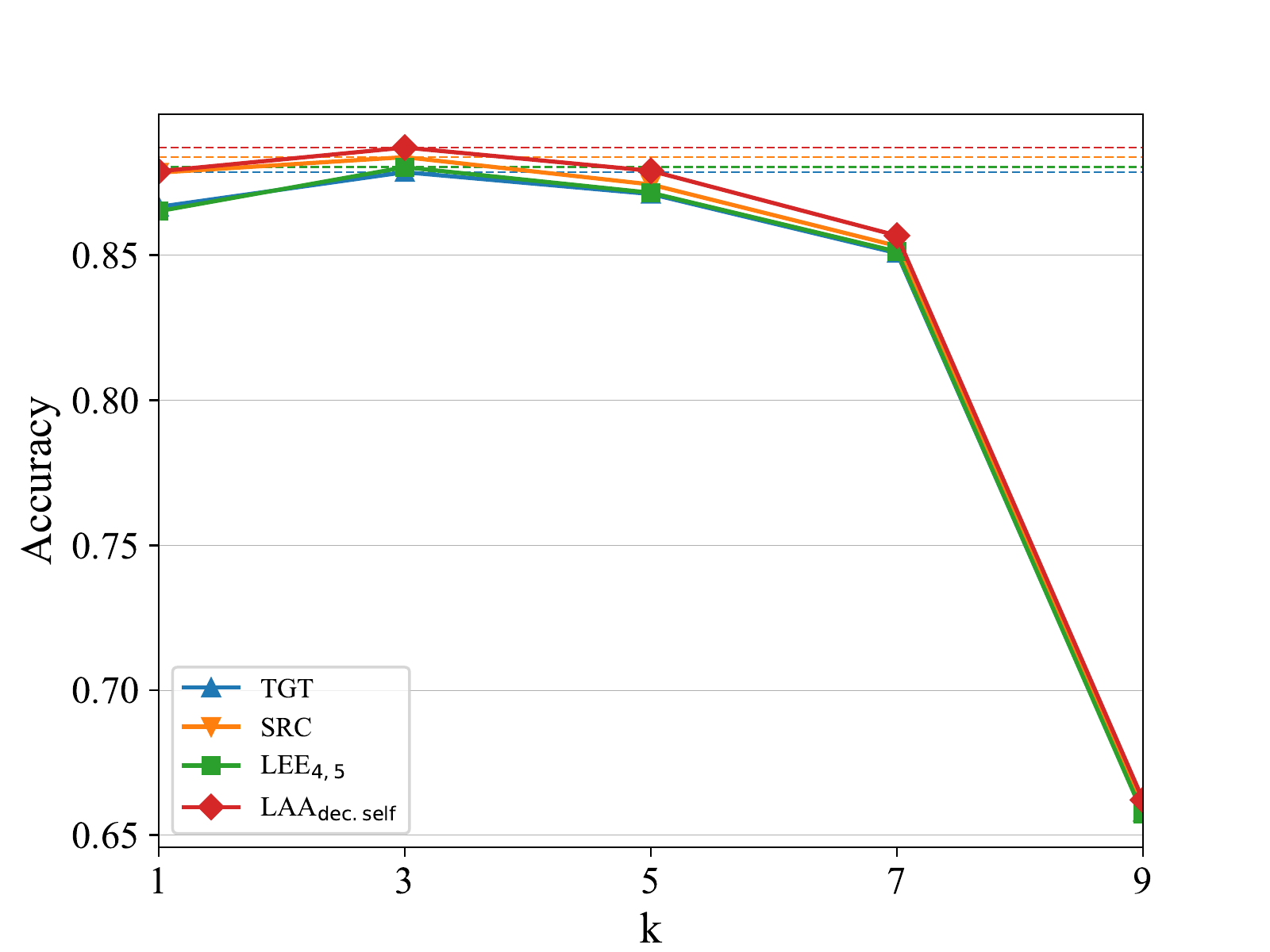}  
            \caption{Phonetic inventory}
        \end{subfigure}
    \caption{Prediction accuracy on syntax, phonology and phonetic inventory features using the language embeddings learned by Token$_{\mathrm{tgt}}$, Token$_{\mathrm{src}}$, LEE$_{4,5}$ and LAA$_{\mathrm{dec.self}}$ which are trained on the TED-59 dataset.}\label{typology_results_ted}
\end{figure*}
\subsubsection{Language Embedding Embodiment}
We selected language embeddings learned by Token$_{\mathrm{src}}$, Token$_{\mathrm{tgt}}$ and LEE$_{4,5}$ to perform linguistic typology prediction given that Token$_{\mathrm{src}}$ and Token$_{\mathrm{tgt}}$ are the dominant practice in MNMT and LEE$_{4,5}$ achieves fairly good performance over all supervised translation directions across the two datasets. We present the experimental results in Table~\ref{lee_typology_results_appendix_table} (Appendix~\ref{typology_results_for_lee}). Optimal $k$s for Token$_{\mathrm{src}}$, Token$_{\mathrm{tgt}}$ and LEE$_{4,5}$ are not always consistent and it is difficult to conclude the most superior one. We therefore adopt the maximum accuracy under different settings of $k$ as the evaluation metric. We observe that there is no significant performance gap except inferring the phonology features with the language embeddings trained on the TED-59 dataset (\circled{9}, Table~\ref{lee_typology_results_appendix_table} in Appendix~\ref{typology_results_for_lee}).

\subsubsection{Language Aware Multi-Head Attention}
Figure~\ref{typology_results_ted} and \ref{typology_results_opus} (the latter shown in Appendix~\ref{typology_results_opus_section}) show the prediction results of using the language representation learned by LAA for linguistic typology probing. We observe that the language representation learned in LAA achieves the best accuracy on syntax feature inference across the two datasets, which demonstrates its generalization ability to some extent. For phonology and phonetic inventory feature inference, there are no significant differences among the language representations learned in various MNMT models. We hypothesize that typological properties encoded in language representations are relied on the task that learns language representations. The translation task can force MNMT models to learn syntactic information to accommodate syntactic divergences across languages for better translation. Additionally, the superior performance of LAA on many-to-many translation and linguistic typology prediction suggests that the language representations learned in LAA can benefit multilingual translation.

\subsection{Effect of the Increased Parameters Introduced by LAA}
\begin{table*}[t]
\centering
    \resizebox{\textwidth}{!}{
        \begin{tabular}{c|c|l|c|ll|ll|ll|lll}
        \toprule
            \multicolumn{1}{c}{\multirow{2}{*}{ID}} & \multicolumn{1}{c}{\multirow{2}{*}{Dataset}} & \multicolumn{1}{c}{\multirow{2}{*}{Model}} & \multicolumn{1}{c}{\multirow{2}{*}{\#Param}} & \multicolumn{2}{c|}{En $\rightarrow$ XX} &  \multicolumn{2}{c|}{XX $\rightarrow$ En} & \multicolumn{2}{c|}{All} & \multicolumn{3}{c}{Zero-shot} \\
        \cmidrule{5-13}
            \multicolumn{1}{c}{} & \multicolumn{1}{c}{} & \multicolumn{1}{c}{} & \multicolumn{1}{c}{} & BLEU & WR & BLEU & WR & BLEU & WR & BLEU & LangAcc & WR \\
        \midrule
            \circled{1} & TED-59 & Token$_{\mathrm{tgt}}$ & 77M & 20.74 & \textit{ref} & 24.08 & \textit{ref} & 22.41 & \textit{ref} & 2.42 & 37.62 & \textit{ref} \\
            \circled{2} & TED-59 & Token$_{\mathrm{src}}$ & 77M & 21.24 & \textbf{91.38} & 23.77 & 15.52 & 22.50 & 53.45 & \textbf{10.50} & 71.82 & 98.91 \\
            \circled{3} & TED-59 & LAA$_{\mathrm{dec.self}}$ & 92M & \textbf{21.29} & 84.48 & 25.14 & 96.55 & 23.21 & 90.52 & 8.94 & 74.68 & 98.58 \\
            \circled{4} & TED-59 & LAA$_{\mathrm{dec.self}}$ + LEE$_{4,5}$ & 92M & 21.28 & 87.93 & 25.29 & \textbf{100.00} & 23.28 & \textbf{93.97} & 9.48 & \textbf{74.69} & \textbf{99.09} \\
            \circled{5} & TED-59 & LAA$_{\mathrm{dec.self}}^{R}$ + Token$_{\mathrm{src}}$ & 92M & 20.74$\pm$0.02 & 44.14$\pm$3.78 & 25.94$\pm$0.01 & 98.28$\pm$0.00 & \textbf{23.34$\pm$0.01} & 71.21$\pm$1.89 & 7.08$\pm$0.00 & 55.73$\pm$0.01 & 95.50$\pm$0.09 \\
            \circled{6} & TED-59 & LAA$_{\mathrm{dec.self}}^{R}$ + Token$_{\mathrm{tgt}}$ & 92M & 20.18$\pm$0.01 & 10.34$\pm$1.22 & \textbf{26.25$\pm$0.02} & 98.28$\pm$0.00 & 23.22$\pm$0.01 & 54.31$\pm$0.61 & 0.92$\pm$0.00 & 6.76$\pm$0.00 & 2.75$\pm$0.09 \\
        \midrule
            \circled{7} & OPUS-100 & Token$_{\mathrm{tgt}}$ & 77M & 23.46 & \textit{ref} & 29.49 & \textit{ref} & 26.47 & \textit{ref} & 5.81 & 55.92 & \textit{ref} \\
            \circled{8} & OPUS-100 & Token$_{\mathrm{src}}$ & 77M & 24.04 & 81.91 & 28.74 & 8.51 & 26.39 & 45.21 & 4.06 & 34.53 & 23.33 \\
            \circled{9} & OPUS-100 & LAA$_{\mathrm{dec.self}}$ & 103M & \textbf{24.42} & \textbf{90.43} & 29.78 & \textbf{79.79} & 27.10 & \textbf{85.11} & 11.23 & 81.53 & \textbf{100.00} \\
            \circled{\small 10} & OPUS-100 & LAA$_{\mathrm{dec.self}}$ + LEE$_{4,5}$ & 103M & 24.41 & 88.30 & \textbf{29.83} & 76.60 & \textbf{27.12} & 82.45 & \textbf{11.31} & \textbf{81.70} & \textbf{100.00} \\
            \circled{\small 11} & OPUS-100 & LAA$_{\mathrm{dec.self}}^{R}$ + Token$_{\mathrm{src}}$ & 103M & 23.65$\pm$0.02 & 64.68$\pm$3.79 & 28.61$\pm$0.01 & 12.98$\pm$0.89 & 26.13$\pm$0.01 & 38.83$\pm$1.50 & 4.18$\pm$0.01 & 39.94$\pm$0.04 & 20.00$\pm$0.00 \\
            \circled{\small 12}& OPUS-100 & LAA$_{\mathrm{dec.self}}^{R}$ + Token$_{\mathrm{tgt}}$ & 103M & 22.80$\pm$0.02 & 16.81$\pm$1.39 & 29.27$\pm$0.01 & 22.13$\pm$2.43 & 26.04$\pm$0.01 & 19.47$\pm$1.22 & 6.16$\pm$0.01 & 68.00$\pm$0.11 & 50.00$\pm$2.36 \\
        \bottomrule
        \end{tabular}
    }
\caption{Experiment results of LAA$_{\mathrm{dec.self}}^{R}$ on the two datasets. The superscript "R" denotes that the language-specific matrices in LAA$_{\mathrm{dec.self}}$ are stochastically activated during training and inference. We adopted Token$_{\mathrm{tgt}}$ or Token$_{\mathrm{src}}$ with LAA$_{\mathrm{dec.self}}^{R}$ to guide the MNMT models into the right translation directions. The results of \circled{1}\circled{2}\circled{7}\circled{8} are from Table~\ref{pre_mnmt_results}. The results of \circled{3}\circled{4}\circled{9}\circled{\small 10} are from Table~\ref{mnmt_results}.}\label{mnmt_results_random}
\end{table*}
We carried out experiments to study the effect of the increased parameters introduced by LAA. Specifically, we removed the language-specific properties of the introduced trainable matrices in LAA by randomly activating them with equal probabilities during training and inference. We denote the MNMT models with the random matrices as LAA$^{R}$. As these matrices become random, there is no explicit signal to guide the translation into the desired target languages. Hence we followed the prepending token strategies of Token$_{\mathrm{tgt}}$ and Token$_{\mathrm{src}}$ to navigate the translation directions.\footnote{We also attempted to make these stochastic matrices language-specific by gradually increasing the sampling probability on a specific matrix for each language during training but results are not satisfactory.} We incorporated the stochastic matrices into the self-attention modules of the decoder since LAA$_{\mathrm{dec.self}}$ achieves superior performance on both supervised and zero-shot translation directions as shown in Table~\ref{mnmt_results} and~\ref{mnmt_results_no_over_sample}. For efficiency, the best checkpoint was selected according to the average BLEU on the validation sets with only one random seed. We evaluated models on the test sets with the best checkpoint and report the mean and standard deviation of the BLEU score with 5 different random seeds. Results with the temperature-based sampling strategy ($T=5$) on the two datasets are shown in Table~\ref{mnmt_results_random}. We also conducted experiments on the raw data distribution and results are presented in Table~\ref{mnmt_results_random_no_oversample} of Appendix~\ref{mnmt_results_random_no_oversample_section}.

We observe that the performance of LAA$_{\mathrm{dec.self}}^{R}$ with different random seeds is stable as the standard deviations on various metrics are small. Additionally, the performance gap between LAA$_{\mathrm{dec.self}}$ and LAA$_{\mathrm{dec.self}}^{R}$ is small on supervised translation directions on the TED-59 dataset. Despite that, LAA$_{\mathrm{dec.self}}$ outperforms LAA$_{\mathrm{dec.self}}^{R}$ by a large margin on the OPUS-100 dataset. We hypothesize that the TED-59 dataset covers more restrict domains than the OPUS-100 dataset sampled from OPUS collection \citep{DBLP:conf/lrec/Tiedemann12},\footnote{There are various corpora in the OPUS collection, such as Wikipedia corpus, Bible corpus, UN corpus and TED corpus, etc. As the OPUS-100 dataset is constructed by randomly sampling sentence pairs from the corpora of OPUS collection for each language pair, it may cover more diverse domains.} which may benefit LAA$_{\mathrm{dec.self}}^{R}$ as the knowledge from similar domains can be easily transferred across the stochastic matrices during training. Furthermore, LAA$_{\mathrm{dec.self}}^{R}$ consistently lags behind LAA$_{\mathrm{dec.self}}$ on zero-shot translation directions on the two datasets. \textbf{Given the same number of parameters in LAA$_{\mathrm{dec.self}}^{R}$ and LAA$_{\mathrm{dec.self}}$, we can conclude that the performance improvement of LAA$_{\mathrm{dec.self}}$ is not only due to the parameter increasement.}

\section{Conclusion}
To improve massively MNMT, we have presented two approaches to learning informative language representations, language embedding embodiment and language-aware multi-head attention. We find that the ways to inject target language information into MNMT models have a significant impact on the translation performance, especially on zero-shot translation. We validate the effectiveness of the two approaches on two public datasets. We probe the typological features encoded in language representations learned by LEE and LAA through linguistic typology feature prediction. The superior prediction performance of the matrix-based language representations learned by LAA on syntax features demonstrates its informativeness.

\section*{Acknowledgements}
The present research was supported by the Key Research and Development Program of Yunnan Province (Grant No. 202203AA080004-2). We would like to thank the anonymous reviewers for their insightful comments.

\clearpage

\bibliography{acl_latex}

\begin{thebibliography}{62}
\expandafter\ifx\csname natexlab\endcsname\relax\def\natexlab#1{#1}\fi

\bibitem[{Aharoni et~al.(2019)Aharoni, Johnson, and
  Firat}]{DBLP:conf/naacl/AharoniJF19}
Roee Aharoni, Melvin Johnson, and Orhan Firat. 2019.
\newblock \href {https://doi.org/10.18653/v1/n19-1388} {Massively multilingual
  neural machine translation}.
\newblock In \emph{Proceedings of the 2019 Conference of the North American
  Chapter of the Association for Computational Linguistics: Human Language
  Technologies, {NAACL-HLT} 2019, Minneapolis, MN, USA, June 2-7, 2019, Volume
  1 (Long and Short Papers)}, pages 3874--3884. Association for Computational
  Linguistics.

\bibitem[{Arivazhagan et~al.(2019{\natexlab{a}})Arivazhagan, Bapna, Firat,
  Aharoni, Johnson, and Macherey}]{DBLP:journals/corr/abs-1903-07091}
Naveen Arivazhagan, Ankur Bapna, Orhan Firat, Roee Aharoni, Melvin Johnson, and
  Wolfgang Macherey. 2019{\natexlab{a}}.
\newblock \href {http://arxiv.org/abs/1903.07091} {The missing ingredient in
  zero-shot neural machine translation}.
\newblock \emph{CoRR}, abs/1903.07091.

\bibitem[{Arivazhagan et~al.(2019{\natexlab{b}})Arivazhagan, Bapna, Firat,
  Lepikhin, Johnson, Krikun, Chen, Cao, Foster, Cherry, Macherey, Chen, and
  Wu}]{DBLP:journals/corr/abs-1907-05019}
Naveen Arivazhagan, Ankur Bapna, Orhan Firat, Dmitry Lepikhin, Melvin Johnson,
  Maxim Krikun, Mia~Xu Chen, Yuan Cao, George~F. Foster, Colin Cherry, Wolfgang
  Macherey, Zhifeng Chen, and Yonghui Wu. 2019{\natexlab{b}}.
\newblock \href {http://arxiv.org/abs/1907.05019} {Massively multilingual
  neural machine translation in the wild: Findings and challenges}.
\newblock \emph{CoRR}, abs/1907.05019.

\bibitem[{Bapna and Firat(2019)}]{DBLP:conf/emnlp/BapnaF19}
Ankur Bapna and Orhan Firat. 2019.
\newblock \href {https://doi.org/10.18653/v1/D19-1165} {Simple, scalable
  adaptation for neural machine translation}.
\newblock In \emph{Proceedings of the 2019 Conference on Empirical Methods in
  Natural Language Processing and the 9th International Joint Conference on
  Natural Language Processing, {EMNLP-IJCNLP} 2019, Hong Kong, China, November
  3-7, 2019}, pages 1538--1548. Association for Computational Linguistics.

\bibitem[{Bjerva and Augenstein(2018)}]{DBLP:conf/naacl/BjervaA18}
Johannes Bjerva and Isabelle Augenstein. 2018.
\newblock \href {https://doi.org/10.18653/v1/n18-1083} {From phonology to
  syntax: Unsupervised linguistic typology at different levels with language
  embeddings}.
\newblock In \emph{Proceedings of the 2018 Conference of the North American
  Chapter of the Association for Computational Linguistics: Human Language
  Technologies, {NAACL-HLT} 2018, New Orleans, Louisiana, USA, June 1-6, 2018,
  Volume 1 (Long Papers)}, pages 907--916. Association for Computational
  Linguistics.

\bibitem[{Bjerva et~al.(2019)Bjerva, {\"{O}}stling, Veiga, Tiedemann, and
  Augenstein}]{DBLP:journals/coling/BjervaOVTA19}
Johannes Bjerva, Robert {\"{O}}stling, Maria~Han Veiga, J{\"{o}}rg Tiedemann,
  and Isabelle Augenstein. 2019.
\newblock \href {https://doi.org/10.1162/coli\_a\_00351} {What do language
  representations really represent?}
\newblock \emph{Comput. Linguistics}, 45(2):381--389.

\bibitem[{Blackwood et~al.(2018)Blackwood, Ballesteros, and
  Ward}]{DBLP:conf/coling/BlackwoodBW18}
Graeme~W. Blackwood, Miguel Ballesteros, and Todd Ward. 2018.
\newblock \href {https://aclanthology.org/C18-1263/} {Multilingual neural
  machine translation with task-specific attention}.
\newblock In \emph{Proceedings of the 27th International Conference on
  Computational Linguistics, {COLING} 2018, Santa Fe, New Mexico, USA, August
  20-26, 2018}, pages 3112--3122. Association for Computational Linguistics.

\bibitem[{Collobert et~al.(2011)Collobert, Weston, Bottou, Karlen, Kavukcuoglu,
  and Kuksa}]{DBLP:journals/jmlr/CollobertWBKKK11}
Ronan Collobert, Jason Weston, L{\'{e}}on Bottou, Michael Karlen, Koray
  Kavukcuoglu, and Pavel~P. Kuksa. 2011.
\newblock \href {http://dl.acm.org/citation.cfm?id=2078186} {Natural language
  processing (almost) from scratch}.
\newblock \emph{J. Mach. Learn. Res.}, 12:2493--2537.

\bibitem[{Conneau and Lample(2019)}]{DBLP:conf/nips/ConneauL19}
Alexis Conneau and Guillaume Lample. 2019.
\newblock \href
  {https://proceedings.neurips.cc/paper/2019/hash/c04c19c2c2474dbf5f7ac4372c5b9af1-Abstract.html}
  {Cross-lingual language model pretraining}.
\newblock In \emph{Advances in Neural Information Processing Systems 32: Annual
  Conference on Neural Information Processing Systems 2019, NeurIPS 2019,
  December 8-14, 2019, Vancouver, BC, Canada}, pages 7057--7067.

\bibitem[{Dong et~al.(2015)Dong, Wu, He, Yu, and
  Wang}]{DBLP:conf/acl/DongWHYW15}
Daxiang Dong, Hua Wu, Wei He, Dianhai Yu, and Haifeng Wang. 2015.
\newblock \href {https://doi.org/10.3115/v1/p15-1166} {Multi-task learning for
  multiple language translation}.
\newblock In \emph{Proceedings of the 53rd Annual Meeting of the Association
  for Computational Linguistics and the 7th International Joint Conference on
  Natural Language Processing of the Asian Federation of Natural Language
  Processing, {ACL} 2015, July 26-31, 2015, Beijing, China, Volume 1: Long
  Papers}, pages 1723--1732. The Association for Computer Linguistics.

\bibitem[{Dryer and Haspelmath(2013)}]{wals}
Matthew~S. Dryer and Martin Haspelmath, editors. 2013.
\newblock \href {https://wals.info/} {\emph{WALS Online}}.
\newblock Max Planck Institute for Evolutionary Anthropology, Leipzig.

\bibitem[{Fan et~al.(2021)Fan, Bhosale, Schwenk, Ma, El{-}Kishky, Goyal,
  Baines, Celebi, Wenzek, Chaudhary, Goyal, Birch, Liptchinsky, Edunov, Auli,
  and Joulin}]{DBLP:journals/jmlr/FanBSMEGBCWCGBL21}
Angela Fan, Shruti Bhosale, Holger Schwenk, Zhiyi Ma, Ahmed El{-}Kishky,
  Siddharth Goyal, Mandeep Baines, Onur Celebi, Guillaume Wenzek, Vishrav
  Chaudhary, Naman Goyal, Tom Birch, Vitaliy Liptchinsky, Sergey Edunov,
  Michael Auli, and Armand Joulin. 2021.
\newblock \href {http://jmlr.org/papers/v22/20-1307.html} {Beyond
  english-centric multilingual machine translation}.
\newblock \emph{J. Mach. Learn. Res.}, 22:107:1--107:48.

\bibitem[{Fedus et~al.(2021)Fedus, Zoph, and
  Shazeer}]{DBLP:journals/corr/abs-2101-03961}
William Fedus, Barret Zoph, and Noam Shazeer. 2021.
\newblock \href {http://arxiv.org/abs/2101.03961} {Switch transformers: Scaling
  to trillion parameter models with simple and efficient sparsity}.
\newblock \emph{CoRR}, abs/2101.03961.

\bibitem[{Firat et~al.(2016{\natexlab{a}})Firat, Cho, and
  Bengio}]{DBLP:conf/naacl/FiratCB16}
Orhan Firat, Kyunghyun Cho, and Yoshua Bengio. 2016{\natexlab{a}}.
\newblock \href {https://doi.org/10.18653/v1/n16-1101} {Multi-way, multilingual
  neural machine translation with a shared attention mechanism}.
\newblock In \emph{{NAACL} {HLT} 2016, The 2016 Conference of the North
  American Chapter of the Association for Computational Linguistics: Human
  Language Technologies, San Diego California, USA, June 12-17, 2016}, pages
  866--875. The Association for Computational Linguistics.

\bibitem[{Firat et~al.(2016{\natexlab{b}})Firat, Sankaran, Al{-}Onaizan,
  Yarman{-}Vural, and Cho}]{DBLP:conf/emnlp/FiratSAYC16}
Orhan Firat, Baskaran Sankaran, Yaser Al{-}Onaizan, Fatos~T. Yarman{-}Vural,
  and Kyunghyun Cho. 2016{\natexlab{b}}.
\newblock \href {https://doi.org/10.18653/v1/d16-1026} {Zero-resource
  translation with multi-lingual neural machine translation}.
\newblock In \emph{Proceedings of the 2016 Conference on Empirical Methods in
  Natural Language Processing, {EMNLP} 2016, Austin, Texas, USA, November 1-4,
  2016}, pages 268--277. The Association for Computational Linguistics.

\bibitem[{Freitag and Firat(2020)}]{DBLP:conf/wmt/FreitagF20}
Markus Freitag and Orhan Firat. 2020.
\newblock \href {https://aclanthology.org/2020.wmt-1.66/} {Complete
  multilingual neural machine translation}.
\newblock In \emph{Proceedings of the Fifth Conference on Machine Translation,
  WMT@EMNLP 2020, Online, November 19-20, 2020}, pages 550--560. Association
  for Computational Linguistics.

\bibitem[{Gong et~al.(2021)Gong, Li, and
  Genzel}]{DBLP:journals/corr/abs-2104-07358}
Hongyu Gong, Xian Li, and Dmitriy Genzel. 2021.
\newblock \href {http://arxiv.org/abs/2104.07358} {Adaptive sparse transformer
  for multilingual translation}.
\newblock \emph{CoRR}, abs/2104.07358.

\bibitem[{Gu et~al.(2018)Gu, Hassan, Devlin, and Li}]{DBLP:conf/naacl/GuHDL18}
Jiatao Gu, Hany Hassan, Jacob Devlin, and Victor O.~K. Li. 2018.
\newblock \href {https://doi.org/10.18653/v1/n18-1032} {Universal neural
  machine translation for extremely low resource languages}.
\newblock In \emph{Proceedings of the 2018 Conference of the North American
  Chapter of the Association for Computational Linguistics: Human Language
  Technologies, {NAACL-HLT} 2018, New Orleans, Louisiana, USA, June 1-6, 2018,
  Volume 1 (Long Papers)}, pages 344--354. Association for Computational
  Linguistics.

\bibitem[{Gu et~al.(2019)Gu, Wang, Cho, and Li}]{DBLP:conf/acl/GuWCL19}
Jiatao Gu, Yong Wang, Kyunghyun Cho, and Victor O.~K. Li. 2019.
\newblock \href {https://doi.org/10.18653/v1/p19-1121} {Improved zero-shot
  neural machine translation via ignoring spurious correlations}.
\newblock In \emph{Proceedings of the 57th Conference of the Association for
  Computational Linguistics, {ACL} 2019, Florence, Italy, July 28- August 2,
  2019, Volume 1: Long Papers}, pages 1258--1268. Association for Computational
  Linguistics.

\bibitem[{Ha et~al.(2016)Ha, Niehues, and Waibel}]{DBLP:conf/iwslt/HaNW16}
Thanh{-}Le Ha, Jan Niehues, and Alex Waibel. 2016.
\newblock \href {https://aclanthology.org/2016.iwslt-1.6} {Toward multilingual
  neural machine translation with universal encoder and decoder}.
\newblock In \emph{Proceedings of the 13th International Conference on Spoken
  Language Translation, {IWSLT} 2016, Seattle, WA, USA, December 8-9, 2016}.
  International Workshop on Spoken Language Translation.

\bibitem[{Hammarström et~al.(2021)Hammarström, Forkel, Haspelmath, and
  Bank}]{glottolog}
Harald Hammarström, Robert Forkel, Martin Haspelmath, and Sebastian Bank.
  2021.
\newblock \href {https://doi.org/10.5281/zenodo.4761960} {\emph{Glottolog
  4.4}}.
\newblock Leipzig.

\bibitem[{Johnson et~al.(2017)Johnson, Schuster, Le, Krikun, Wu, Chen, Thorat,
  Vi{\'{e}}gas, Wattenberg, Corrado, Hughes, and
  Dean}]{DBLP:journals/tacl/JohnsonSLKWCTVW17}
Melvin Johnson, Mike Schuster, Quoc~V. Le, Maxim Krikun, Yonghui Wu, Zhifeng
  Chen, Nikhil Thorat, Fernanda~B. Vi{\'{e}}gas, Martin Wattenberg, Greg
  Corrado, Macduff Hughes, and Jeffrey Dean. 2017.
\newblock \href {https://transacl.org/ojs/index.php/tacl/article/view/1081}
  {Google's multilingual neural machine translation system: Enabling zero-shot
  translation}.
\newblock \emph{Trans. Assoc. Comput. Linguistics}, 5:339--351.

\bibitem[{Kingma and Ba(2015)}]{DBLP:journals/corr/KingmaB14}
Diederik~P. Kingma and Jimmy Ba. 2015.
\newblock \href {http://arxiv.org/abs/1412.6980} {Adam: {A} method for
  stochastic optimization}.
\newblock In \emph{3rd International Conference on Learning Representations,
  {ICLR} 2015, San Diego, CA, USA, May 7-9, 2015, Conference Track
  Proceedings}.

\bibitem[{Kong et~al.(2021)Kong, Renduchintala, Cross, Tang, Gu, and
  Li}]{DBLP:conf/eacl/KongRCTGL21}
Xiang Kong, Adithya Renduchintala, James Cross, Yuqing Tang, Jiatao Gu, and
  Xian Li. 2021.
\newblock \href {https://aclanthology.org/2021.eacl-main.138/} {Multilingual
  neural machine translation with deep encoder and multiple shallow decoders}.
\newblock In \emph{Proceedings of the 16th Conference of the European Chapter
  of the Association for Computational Linguistics: Main Volume, {EACL} 2021,
  Online, April 19 - 23, 2021}, pages 1613--1624. Association for Computational
  Linguistics.

\bibitem[{Kudo and Richardson(2018)}]{DBLP:conf/emnlp/KudoR18}
Taku Kudo and John Richardson. 2018.
\newblock \href {https://doi.org/10.18653/v1/d18-2012} {Sentencepiece: {A}
  simple and language independent subword tokenizer and detokenizer for neural
  text processing}.
\newblock In \emph{Proceedings of the 2018 Conference on Empirical Methods in
  Natural Language Processing, {EMNLP} 2018: System Demonstrations, Brussels,
  Belgium, October 31 - November 4, 2018}, pages 66--71. Association for
  Computational Linguistics.

\bibitem[{Lin et~al.(2020)Lin, Pan, Wang, Qiu, Feng, Zhou, and
  Li}]{DBLP:conf/emnlp/LinPWQFZL20}
Zehui Lin, Xiao Pan, Mingxuan Wang, Xipeng Qiu, Jiangtao Feng, Hao Zhou, and
  Lei Li. 2020.
\newblock \href {https://doi.org/10.18653/v1/2020.emnlp-main.210} {Pre-training
  multilingual neural machine translation by leveraging alignment information}.
\newblock In \emph{Proceedings of the 2020 Conference on Empirical Methods in
  Natural Language Processing, {EMNLP} 2020, Online, November 16-20, 2020},
  pages 2649--2663. Association for Computational Linguistics.

\bibitem[{Lin et~al.(2021)Lin, Wu, Wang, and Li}]{DBLP:conf/acl/LinWWL20}
Zehui Lin, Liwei Wu, Mingxuan Wang, and Lei Li. 2021.
\newblock \href {https://doi.org/10.18653/v1/2021.acl-long.25} {Learning
  language specific sub-network for multilingual machine translation}.
\newblock In \emph{Proceedings of the 59th Annual Meeting of the Association
  for Computational Linguistics and the 11th International Joint Conference on
  Natural Language Processing, {ACL/IJCNLP} 2021, (Volume 1: Long Papers),
  Virtual Event, August 1-6, 2021}, pages 293--305. Association for
  Computational Linguistics.

\bibitem[{Littell et~al.(2017)Littell, Mortensen, Lin, Kairis, Turner, and
  Levin}]{DBLP:conf/eacl/LevinLMLKT17}
Patrick Littell, David~R. Mortensen, Ke~Lin, Katherine Kairis, Carlisle Turner,
  and Lori~S. Levin. 2017.
\newblock \href {https://doi.org/10.18653/v1/e17-2002} {{URIEL} and lang2vec:
  Representing languages as typological, geographical, and phylogenetic
  vectors}.
\newblock In \emph{Proceedings of the 15th Conference of the European Chapter
  of the Association for Computational Linguistics, {EACL} 2017, Valencia,
  Spain, April 3-7, 2017, Volume 2: Short Papers}, pages 8--14. Association for
  Computational Linguistics.

\bibitem[{Luong et~al.(2016)Luong, Le, Sutskever, Vinyals, and
  Kaiser}]{DBLP:journals/corr/LuongLSVK15}
Minh{-}Thang Luong, Quoc~V. Le, Ilya Sutskever, Oriol Vinyals, and Lukasz
  Kaiser. 2016.
\newblock \href {http://arxiv.org/abs/1511.06114} {Multi-task sequence to
  sequence learning}.
\newblock In \emph{4th International Conference on Learning Representations,
  {ICLR} 2016, San Juan, Puerto Rico, May 2-4, 2016, Conference Track
  Proceedings}.

\bibitem[{Malaviya et~al.(2017)Malaviya, Neubig, and
  Littell}]{DBLP:conf/emnlp/MalaviyaNL17}
Chaitanya Malaviya, Graham Neubig, and Patrick Littell. 2017.
\newblock \href {https://doi.org/10.18653/v1/d17-1268} {Learning language
  representations for typology prediction}.
\newblock In \emph{Proceedings of the 2017 Conference on Empirical Methods in
  Natural Language Processing, {EMNLP} 2017, Copenhagen, Denmark, September
  9-11, 2017}, pages 2529--2535. Association for Computational Linguistics.

\bibitem[{Moran and McCloy(2019)}]{phoible}
Steven Moran and Daniel McCloy, editors. 2019.
\newblock \href {https://phoible.org/} {\emph{PHOIBLE 2.0}}.
\newblock Max Planck Institute for the Science of Human History, Jena.

\bibitem[{Neubig and Hu(2018)}]{DBLP:conf/emnlp/NeubigH18}
Graham Neubig and Junjie Hu. 2018.
\newblock \href {https://doi.org/10.18653/v1/d18-1103} {Rapid adaptation of
  neural machine translation to new languages}.
\newblock In \emph{Proceedings of the 2018 Conference on Empirical Methods in
  Natural Language Processing, Brussels, Belgium, October 31 - November 4,
  2018}, pages 875--880. Association for Computational Linguistics.

\bibitem[{Oncevay et~al.(2020)Oncevay, Haddow, and
  Birch}]{DBLP:conf/emnlp/OncevayHB20}
Arturo Oncevay, Barry Haddow, and Alexandra Birch. 2020.
\newblock \href {https://doi.org/10.18653/v1/2020.emnlp-main.187} {Bridging
  linguistic typology and multilingual machine translation with multi-view
  language representations}.
\newblock In \emph{Proceedings of the 2020 Conference on Empirical Methods in
  Natural Language Processing, {EMNLP} 2020, Online, November 16-20, 2020},
  pages 2391--2406. Association for Computational Linguistics.

\bibitem[{{\"{O}}stling and Tiedemann(2017)}]{DBLP:conf/eacl/TiedemannO17}
Robert {\"{O}}stling and J{\"{o}}rg Tiedemann. 2017.
\newblock \href {https://doi.org/10.18653/v1/e17-2102} {Continuous
  multilinguality with language vectors}.
\newblock In \emph{Proceedings of the 15th Conference of the European Chapter
  of the Association for Computational Linguistics, {EACL} 2017, Valencia,
  Spain, April 3-7, 2017, Volume 2: Short Papers}, pages 644--649. Association
  for Computational Linguistics.

\bibitem[{Ott et~al.(2019)Ott, Edunov, Baevski, Fan, Gross, Ng, Grangier, and
  Auli}]{DBLP:conf/naacl/OttEBFGNGA19}
Myle Ott, Sergey Edunov, Alexei Baevski, Angela Fan, Sam Gross, Nathan Ng,
  David Grangier, and Michael Auli. 2019.
\newblock \href {https://doi.org/10.18653/v1/n19-4009} {fairseq: {A} fast,
  extensible toolkit for sequence modeling}.
\newblock In \emph{Proceedings of the 2019 Conference of the North American
  Chapter of the Association for Computational Linguistics: Human Language
  Technologies, {NAACL-HLT} 2019, Minneapolis, MN, USA, June 2-7, 2019,
  Demonstrations}, pages 48--53. Association for Computational Linguistics.

\bibitem[{Pan et~al.(2021)Pan, Wang, Wu, and Li}]{DBLP:conf/acl/PanWWL20}
Xiao Pan, Mingxuan Wang, Liwei Wu, and Lei Li. 2021.
\newblock \href {https://doi.org/10.18653/v1/2021.acl-long.21} {Contrastive
  learning for many-to-many multilingual neural machine translation}.
\newblock In \emph{Proceedings of the 59th Annual Meeting of the Association
  for Computational Linguistics and the 11th International Joint Conference on
  Natural Language Processing, {ACL/IJCNLP} 2021, (Volume 1: Long Papers),
  Virtual Event, August 1-6, 2021}, pages 244--258. Association for
  Computational Linguistics.

\bibitem[{Pham et~al.(2019)Pham, Niehues, Ha, and
  Waibel}]{DBLP:conf/wmt/PhamNHW19}
Ngoc{-}Quan Pham, Jan Niehues, Thanh{-}Le Ha, and Alexander Waibel. 2019.
\newblock \href {https://doi.org/10.18653/v1/w19-5202} {Improving zero-shot
  translation with language-independent constraints}.
\newblock In \emph{Proceedings of the Fourth Conference on Machine Translation,
  {WMT} 2019, Florence, Italy, August 1-2, 2019 - Volume 1: Research Papers},
  pages 13--23. Association for Computational Linguistics.

\bibitem[{Philip et~al.(2020)Philip, Berard, Gall{\'{e}}, and
  Besacier}]{DBLP:conf/emnlp/PhilipBGB20}
Jerin Philip, Alexandre Berard, Matthias Gall{\'{e}}, and Laurent Besacier.
  2020.
\newblock \href {https://doi.org/10.18653/v1/2020.emnlp-main.361} {Monolingual
  adapters for zero-shot neural machine translation}.
\newblock In \emph{Proceedings of the 2020 Conference on Empirical Methods in
  Natural Language Processing, {EMNLP} 2020, Online, November 16-20, 2020},
  pages 4465--4470. Association for Computational Linguistics.

\bibitem[{Post(2018)}]{DBLP:conf/wmt/Post18}
Matt Post. 2018.
\newblock \href {https://doi.org/10.18653/v1/w18-6319} {A call for clarity in
  reporting {BLEU} scores}.
\newblock In \emph{Proceedings of the Third Conference on Machine Translation:
  Research Papers, {WMT} 2018, Belgium, Brussels, October 31 - November 1,
  2018}, pages 186--191. Association for Computational Linguistics.

\bibitem[{Qi et~al.(2018)Qi, Sachan, Felix, Padmanabhan, and
  Neubig}]{DBLP:conf/naacl/QiSFPN18}
Ye~Qi, Devendra~Singh Sachan, Matthieu Felix, Sarguna Padmanabhan, and Graham
  Neubig. 2018.
\newblock \href {https://doi.org/10.18653/v1/n18-2084} {When and why are
  pre-trained word embeddings useful for neural machine translation?}
\newblock In \emph{Proceedings of the 2018 Conference of the North American
  Chapter of the Association for Computational Linguistics: Human Language
  Technologies, NAACL-HLT, New Orleans, Louisiana, USA, June 1-6, 2018, Volume
  2 (Short Papers)}, pages 529--535. Association for Computational Linguistics.

\bibitem[{Raganato et~al.(2021)Raganato, V{\'{a}}zquez, Creutz, and
  Tiedemann}]{DBLP:conf/emnlp/RaganatoVCT21}
Alessandro Raganato, Ra{\'{u}}l V{\'{a}}zquez, Mathias Creutz, and J{\"{o}}rg
  Tiedemann. 2021.
\newblock \href {https://doi.org/10.18653/v1/2021.emnlp-main.664} {An empirical
  investigation of word alignment supervision for zero-shot multilingual neural
  machine translation}.
\newblock In \emph{Proceedings of the 2021 Conference on Empirical Methods in
  Natural Language Processing, {EMNLP} 2021, Virtual Event / Punta Cana,
  Dominican Republic, 7-11 November, 2021}, pages 8449--8456. Association for
  Computational Linguistics.

\bibitem[{Rios et~al.(2020)Rios, M{\"{u}}ller, and
  Sennrich}]{DBLP:conf/wmt/RiosMS20}
Annette Rios, Mathias M{\"{u}}ller, and Rico Sennrich. 2020.
\newblock \href {https://aclanthology.org/2020.wmt-1.64/} {Subword segmentation
  and a single bridge language affect zero-shot neural machine translation}.
\newblock In \emph{Proceedings of the Fifth Conference on Machine Translation,
  WMT@EMNLP 2020, Online, November 19-20, 2020}, pages 528--537. Association
  for Computational Linguistics.

\bibitem[{Sachan and Neubig(2018)}]{DBLP:conf/wmt/SachanN18}
Devendra~Singh Sachan and Graham Neubig. 2018.
\newblock \href {https://doi.org/10.18653/v1/w18-6327} {Parameter sharing
  methods for multilingual self-attentional translation models}.
\newblock In \emph{Proceedings of the Third Conference on Machine Translation:
  Research Papers, {WMT} 2018, Belgium, Brussels, October 31 - November 1,
  2018}, pages 261--271. Association for Computational Linguistics.

\bibitem[{Sennrich et~al.(2016)Sennrich, Haddow, and
  Birch}]{DBLP:conf/acl/SennrichHB16a}
Rico Sennrich, Barry Haddow, and Alexandra Birch. 2016.
\newblock \href {https://doi.org/10.18653/v1/p16-1162} {Neural machine
  translation of rare words with subword units}.
\newblock In \emph{Proceedings of the 54th Annual Meeting of the Association
  for Computational Linguistics, {ACL} 2016, August 7-12, 2016, Berlin,
  Germany, Volume 1: Long Papers}. The Association for Computer Linguistics.

\bibitem[{Tan et~al.(2019)Tan, Chen, He, Xia, Qin, and
  Liu}]{DBLP:conf/emnlp/TanCHXQL19}
Xu~Tan, Jiale Chen, Di~He, Yingce Xia, Tao Qin, and Tie{-}Yan Liu. 2019.
\newblock \href {https://doi.org/10.18653/v1/D19-1089} {Multilingual neural
  machine translation with language clustering}.
\newblock In \emph{Proceedings of the 2019 Conference on Empirical Methods in
  Natural Language Processing and the 9th International Joint Conference on
  Natural Language Processing, {EMNLP-IJCNLP} 2019, Hong Kong, China, November
  3-7, 2019}, pages 963--973. Association for Computational Linguistics.

\bibitem[{Tiedemann(2012)}]{DBLP:conf/lrec/Tiedemann12}
J{\"{o}}rg Tiedemann. 2012.
\newblock \href
  {http://www.lrec-conf.org/proceedings/lrec2012/summaries/463.html} {Parallel
  data, tools and interfaces in {OPUS}}.
\newblock In \emph{Proceedings of the Eighth International Conference on
  Language Resources and Evaluation, {LREC} 2012, Istanbul, Turkey, May 23-25,
  2012}, pages 2214--2218. European Language Resources Association {(ELRA)}.

\bibitem[{Vaswani et~al.(2017)Vaswani, Shazeer, Parmar, Uszkoreit, Jones,
  Gomez, Kaiser, and Polosukhin}]{DBLP:conf/nips/VaswaniSPUJGKP17}
Ashish Vaswani, Noam Shazeer, Niki Parmar, Jakob Uszkoreit, Llion Jones,
  Aidan~N. Gomez, Lukasz Kaiser, and Illia Polosukhin. 2017.
\newblock \href
  {https://proceedings.neurips.cc/paper/2017/hash/3f5ee243547dee91fbd053c1c4a845aa-Abstract.html}
  {Attention is all you need}.
\newblock In \emph{Advances in Neural Information Processing Systems 30: Annual
  Conference on Neural Information Processing Systems 2017, December 4-9, 2017,
  Long Beach, CA, {USA}}, pages 5998--6008.

\bibitem[{V{\'{a}}zquez et~al.(2020)V{\'{a}}zquez, Raganato, Creutz, and
  Tiedemann}]{DBLP:journals/coling/VazquezRCT20}
Ra{\'{u}}l V{\'{a}}zquez, Alessandro Raganato, Mathias Creutz, and J{\"{o}}rg
  Tiedemann. 2020.
\newblock \href {https://doi.org/10.1162/coli\_a\_00377} {A systematic study of
  inner-attention-based sentence representations in multilingual neural machine
  translation}.
\newblock \emph{Comput. Linguistics}, 46(2):387--424.

\bibitem[{Wang et~al.(2022)Wang, Ma, Dong, Huang, Zhang, and
  Wei}]{DBLP:journals/corr/abs-2203-00555}
Hongyu Wang, Shuming Ma, Li~Dong, Shaohan Huang, Dongdong Zhang, and Furu Wei.
  2022.
\newblock \href {https://doi.org/10.48550/arXiv.2203.00555} {Deepnet: Scaling
  transformers to 1, 000 layers}.
\newblock \emph{CoRR}, abs/2203.00555.

\bibitem[{Wang and Zhang(2022)}]{DBLP:conf/aaai/WangZ22}
Qian Wang and Jiajun Zhang. 2022.
\newblock \href {https://ojs.aaai.org/index.php/AAAI/article/view/21396}
  {Parameter differentiation based multilingual neural machine translation}.
\newblock In \emph{Thirty-Sixth {AAAI} Conference on Artificial Intelligence,
  {AAAI} 2022, Thirty-Fourth Conference on Innovative Applications of
  Artificial Intelligence, {IAAI} 2022, The Twelveth Symposium on Educational
  Advances in Artificial Intelligence, {EAAI} 2022 Virtual Event, February 22 -
  March 1, 2022}, pages 11440--11448. {AAAI} Press.

\bibitem[{Wang et~al.(2018)Wang, Zhang, Zhai, Xu, and
  Zong}]{DBLP:conf/emnlp/WangZZXZ18}
Yining Wang, Jiajun Zhang, Feifei Zhai, Jingfang Xu, and Chengqing Zong. 2018.
\newblock \href {https://doi.org/10.18653/v1/d18-1326} {Three strategies to
  improve one-to-many multilingual translation}.
\newblock In \emph{Proceedings of the 2018 Conference on Empirical Methods in
  Natural Language Processing, Brussels, Belgium, October 31 - November 4,
  2018}, pages 2955--2960. Association for Computational Linguistics.

\bibitem[{Wang et~al.(2019)Wang, Zhou, Zhang, Zhai, Xu, and
  Zong}]{DBLP:conf/acl/WangZZZXZ19}
Yining Wang, Long Zhou, Jiajun Zhang, Feifei Zhai, Jingfang Xu, and Chengqing
  Zong. 2019.
\newblock \href {https://doi.org/10.18653/v1/p19-1117} {A compact and
  language-sensitive multilingual translation method}.
\newblock In \emph{Proceedings of the 57th Conference of the Association for
  Computational Linguistics, {ACL} 2019, Florence, Italy, July 28- August 2,
  2019, Volume 1: Long Papers}, pages 1213--1223. Association for Computational
  Linguistics.

\bibitem[{Wang et~al.(2021)Wang, Tsvetkov, Firat, and
  Cao}]{DBLP:conf/iclr/WangTF021}
Zirui Wang, Yulia Tsvetkov, Orhan Firat, and Yuan Cao. 2021.
\newblock \href {https://openreview.net/forum?id=F1vEjWK-lH\_} {Gradient
  vaccine: Investigating and improving multi-task optimization in massively
  multilingual models}.
\newblock In \emph{9th International Conference on Learning Representations,
  {ICLR} 2021, Virtual Event, Austria, May 3-7, 2021}. OpenReview.net.

\bibitem[{Wu et~al.(2021)Wu, Cheng, Wang, and Li}]{DBLP:conf/acl/WuCWL21}
Liwei Wu, Shanbo Cheng, Mingxuan Wang, and Lei Li. 2021.
\newblock \href {https://doi.org/10.18653/v1/2021.findings-acl.264} {Language
  tags matter for zero-shot neural machine translation}.
\newblock In \emph{Findings of the Association for Computational Linguistics:
  {ACL/IJCNLP} 2021, Online Event, August 1-6, 2021}, volume {ACL/IJCNLP} 2021
  of \emph{Findings of {ACL}}, pages 3001--3007. Association for Computational
  Linguistics.

\bibitem[{Xie et~al.(2021)Xie, Feng, Gu, and Yu}]{DBLP:conf/acl/Xie0G020}
Wanying Xie, Yang Feng, Shuhao Gu, and Dong Yu. 2021.
\newblock \href {https://doi.org/10.18653/v1/2021.acl-long.445}
  {Importance-based neuron allocation for multilingual neural machine
  translation}.
\newblock In \emph{Proceedings of the 59th Annual Meeting of the Association
  for Computational Linguistics and the 11th International Joint Conference on
  Natural Language Processing, {ACL/IJCNLP} 2021, (Volume 1: Long Papers),
  Virtual Event, August 1-6, 2021}, pages 5725--5737. Association for
  Computational Linguistics.

\bibitem[{Xu et~al.(2021)Xu, Liu, van Genabith, and
  Xiong}]{DBLP:conf/acl/XuLGX20}
Hongfei Xu, Qiuhui Liu, Josef van Genabith, and Deyi Xiong. 2021.
\newblock \href {https://doi.org/10.18653/v1/2021.acl-short.46} {Modeling
  task-aware {MIMO} cardinality for efficient multilingual neural machine
  translation}.
\newblock In \emph{Proceedings of the 59th Annual Meeting of the Association
  for Computational Linguistics and the 11th International Joint Conference on
  Natural Language Processing, {ACL/IJCNLP} 2021, (Volume 2: Short Papers),
  Virtual Event, August 1-6, 2021}, pages 361--367. Association for
  Computational Linguistics.

\bibitem[{Yang et~al.(2021)Yang, Eriguchi, Muzio, Tadepalli, Lee, and
  Hassan}]{DBLP:conf/emnlp/YangEMTLH21}
Yilin Yang, Akiko Eriguchi, Alexandre Muzio, Prasad Tadepalli, Stefan Lee, and
  Hany Hassan. 2021.
\newblock \href {https://aclanthology.org/2021.emnlp-main.578} {Improving
  multilingual translation by representation and gradient regularization}.
\newblock In \emph{Proceedings of the 2021 Conference on Empirical Methods in
  Natural Language Processing, {EMNLP} 2021, Virtual Event / Punta Cana,
  Dominican Republic, 7-11 November, 2021}, pages 7266--7279. Association for
  Computational Linguistics.

\bibitem[{Yu et~al.(2021)Yu, He, and Sagae}]{DBLP:conf/acl/YuHS20}
Dian Yu, Taiqi He, and Kenji Sagae. 2021.
\newblock \href {https://doi.org/10.18653/v1/2021.acl-long.560} {Language
  embeddings for typology and cross-lingual transfer learning}.
\newblock In \emph{Proceedings of the 59th Annual Meeting of the Association
  for Computational Linguistics and the 11th International Joint Conference on
  Natural Language Processing, {ACL/IJCNLP} 2021, (Volume 1: Long Papers),
  Virtual Event, August 1-6, 2021}, pages 7210--7225. Association for
  Computational Linguistics.

\bibitem[{Zhang et~al.(2021)Zhang, Bapna, Sennrich, and
  Firat}]{DBLP:conf/iclr/ZhangBSF21}
Biao Zhang, Ankur Bapna, Rico Sennrich, and Orhan Firat. 2021.
\newblock \href {https://openreview.net/forum?id=Wj4ODo0uyCF} {Share or not?
  learning to schedule language-specific capacity for multilingual
  translation}.
\newblock In \emph{9th International Conference on Learning Representations,
  {ICLR} 2021, Virtual Event, Austria, May 3-7, 2021}. OpenReview.net.

\bibitem[{Zhang et~al.(2020)Zhang, Williams, Titov, and
  Sennrich}]{DBLP:conf/acl/ZhangWTS20}
Biao Zhang, Philip Williams, Ivan Titov, and Rico Sennrich. 2020.
\newblock \href {https://doi.org/10.18653/v1/2020.acl-main.148} {Improving
  massively multilingual neural machine translation and zero-shot translation}.
\newblock In \emph{Proceedings of the 58th Annual Meeting of the Association
  for Computational Linguistics, {ACL} 2020, Online, July 5-10, 2020}, pages
  1628--1639. Association for Computational Linguistics.

\bibitem[{Zhu et~al.(2021)Zhu, Feng, Zhao, Wang, and
  Li}]{DBLP:conf/emnlp/ZhuFZWL21}
Yaoming Zhu, Jiangtao Feng, Chengqi Zhao, Mingxuan Wang, and Lei Li. 2021.
\newblock \href {https://doi.org/10.18653/v1/2021.findings-emnlp.240}
  {Counter-interference adapter for multilingual machine translation}.
\newblock In \emph{Findings of the Association for Computational Linguistics:
  {EMNLP} 2021, Virtual Event / Punta Cana, Dominican Republic, 16-20 November,
  2021}, pages 2812--2823. Association for Computational Linguistics.

\bibitem[{Zoph and Knight(2016)}]{DBLP:conf/naacl/ZophK16}
Barret Zoph and Kevin Knight. 2016.
\newblock \href {https://doi.org/10.18653/v1/n16-1004} {Multi-source neural
  translation}.
\newblock In \emph{{NAACL} {HLT} 2016, The 2016 Conference of the North
  American Chapter of the Association for Computational Linguistics: Human
  Language Technologies, San Diego California, USA, June 12-17, 2016}, pages
  30--34. The Association for Computational Linguistics.

\end{thebibliography}
\bibliographystyle{acl_natbib}

\clearpage
\appendix

\section{Efficient Implementation of LAA}\label{efficient_implementation_of_laa}
Suppose that the number of target languages is $l$ and the dimensions of $\bm{Q}$, $\bm{K}$, $\bm{V}$ are $b \times n \times d_{model}$, where $b$ is the batch size and $n$ is the sequence length. Since a minibatch usually contains samples from various language pairs during training, each minibatch is required to be split into smaller batches according to the target language so as to compute Eq.~(\ref{equ_3}) and (\ref{equ_5}). And the output of Eq.~(\ref{equ_5}) is required to be reorganized to preserve the original order of samples within the minibatch, which brings extra computational cost and is inefficient for computing on modern parallel hardware like GPU. To avoid this, we maintain a matrix $\widetilde{\bm{W}}\in \mathbb{R}^{l \times d_{model} \times d_{model}}$ which consists of language representations from all languages. We select $\overline{\bm{W}} \in \mathbb{R}^{b \times d_{model} \times d_{model}}$ from $\widetilde{\bm{W}}$, which contains all language representations for samples in the minibatch. This selection can be efficiently executed by the \texttt{Pytorch} toolkit.\footnote{\url{https://pytorch.org/docs/stable/generated/torch.index_select.html?highlight=index_select\#torch.index_select}} To avoid broadcasting\footnote{\url{https://pytorch.org/docs/stable/notes/broadcasting.html?highlight=broadcasting}} $\bm{W}^{Q}$, $\bm{W}^{K}$, $\bm{W}^{V}$, $\bm{W}^{O}$ into larger dimensions when added with $\overline{\bm{W}}$, we reformulate Eq.~(\ref{equ_3}) and (\ref{equ_5}) as follows:
\begin{equation}\label{equ_6}
    \begin{aligned}
    \bm{q} = \bm{Q}\bm{W}^{Q} +  \bm{Q}\overline{\bm{W}}\\
    \bm{k} = \bm{K}\bm{W}^{K} + \bm{K}\overline{\bm{W}}\\
    \bm{v} = \bm{V}\bm{W}^{V} + \bm{V}\overline{\bm{W}}
    \end{aligned}
\end{equation}
\begin{equation}\label{equ_7}
    \bm{Z} = \bm{z}\bm{W}^{O} + \bm{z}\overline{\bm{W}}^{\mathsf{T}}
\end{equation}
Note that we omit the subscript for head index in original formulas as all heads are computed in parallel. As there are always duplicate elements in \{$\bm{Q}$, $\bm{K}$, $\bm{V}$\}, the second term (i.e. \{$\bm{Q}$,$\bm{K}$,$\bm{V}$\}$\overline{\bm{W}}$) in Eq.~(\ref{equ_6}) can be computed first and then cached to avoid redundant computation.

\section{Dataset for Many-to-Many Translation}\label{dataset_for_many_to_many_translation}
We removed the \texttt{\_\_en\_\_} tag prepended to non-English sentences in the original TED-59 dataset.\footnote{An artificial English tag \texttt{\_\_en\_\_} is prepended to every non-English sentence in the raw dataset, which may affect model training and bias BLEU. Nevertheless, previous works on this dataset usually do not elaborate this procedure, which may make our results on this dataset not directly comparable to theirs.} We trained BPE model \citep{DBLP:conf/acl/SennrichHB16a} using SentencePiece \citep{DBLP:conf/emnlp/KudoR18} to get subword units with a joint vocabulary of size 64K.

\section{Model \& Training for Many-to-Many Translation}\label{mnmt_model_training_appendix}
We implement our MNMT models based on \texttt{Fairseq} \citep{DBLP:conf/naacl/OttEBFGNGA19}. We set the dimension of word embeddings and FFN layer to 512/2048. Embeddings were shared for the encoder, decoder and the output projection. To prevent overfitting, we set dropout rate to 0.1 on the OPUS-100 dataset and 0.2 on the TED-59 dataset due to its relatively small data size. We adopted the cross-entropy loss with a label smoothing of 0.1 as the training objective. We used Adam $(\beta_{1} = 0.9, \beta_{2} = 0.98)$ \citep{DBLP:journals/corr/KingmaB14} to optimize model parameters. We varied the learning rate according to the \texttt{inverse\_square\_root} schedule \citep{DBLP:conf/nips/VaswaniSPUJGKP17} with a warm-up step of 4000 and a peak learning rate of 0.0005. We trained all MNMT models for 30 epochs and each minibatch contains a maximum of 4096 tokens. The training data of each epoch is composed of the training datasets from all translation directions. For the OPUS-100 dataset, the number of total training steps for the model with and without oversampling are $\sim$ 933,000 and $\sim$ 617,000 respectively. For the TED-59 dataset, the number of total training steps with and without oversampling are $\sim$ 226,000 and $\sim$ 142,000 respectively.\footnote{To reduce the number of padding tokens in the minibatch, \texttt{Fairseq} sorts the training data by comparing the target sentence length first and then the source sentence length by default. Additionally, the number of tokens in each training sample is computed as the maximum of source and target sentence length, which is used to enforce the minibatch size. As a result, the language tag prepending strategy will affect the number of minibatches. Because of this, we report the approximate number of training steps.} Parallel sentences in the training data sets where the number of subwords on either the source or target side exceeds 100 were removed.

\section{Evaluation of MNMT Model}\label{mnmt_model_evaluation_appendix}
We performed beam search decoding with a beam size of 5 and length penalty of 1.0 during inference. As the TED-59 dataset has already been tokenized, we detokenized reference and system translations with \texttt{sacremoses}\footnote{\url{https://github.com/alvations/sacremoses}} toolkit before computing BLEU. We chose the best checkpoint according to the average BLEU on the validation sets and then evaluated it on the test sets. Considering that there are no validation sets for zero-shot translation directions, we used the checkpoint selected on the supervised translation directions for zero-shot translation. To be comparable with \citep{DBLP:conf/acl/ZhangWTS20}, we employed \texttt{langdetect}\footnote{\url{https://github.com/Mimino666/langdetect}} toolkit for language identification on the OPUS-100 dataset. For TED-59 dataset, we used the \texttt{langid.id} toolkit\footnote{\url{https://github.com/saffsd/langid.py}} instead as it supports more languages than \texttt{langdetect}. We disregarded languages that cannot be detected by \texttt{langid.id} toolkit such as Canadian French (fr-ca), Brazilian Portuguese (pt-br) and Burmese (my).

\section{Dataset for Linguistic Typology Prediction}\label{dataset_for_linguistic_typology_prediction}
URIEL is a typological compendium which accommodates diverse linguistic resources from several typological databases such as WALS \citep{wals}, PHOIBLE \citep{phoible} and Glottolog \citep{glottolog}. We used \texttt{lang2vec}\footnote{\url{https://github.com/antonisa/lang2vec}} library to query URIEL database which provides uniform interface to access various linguistic features.

\section{Prediction Method for Typology Features}\label{prediction_method_for_typology_features}
We inferred typology features of languages from the language representations derived from the trained MNMT model which achieves the best performance on the validation sets. Some previous works train logistic regression classifiers to perform typology prediction \citep{DBLP:conf/emnlp/MalaviyaNL17,DBLP:conf/emnlp/OncevayHB20}. Nevertheless, logistic regression is a parameterized algorithm and the number of its parameters increase with the feature dimensions of input data, which makes it difficult to handle data scarcity with high-dimensional inputs such as matrix. We hence adopted $k$-nearest neighbors approach (k-NN), a non-parametric method, for linguistic typology prediction. We employed cosine similarity as the distance measure for LEE. For the language representations in LAA, we computed the average cosine similarity for all rows in the matrix as the distance metric. We set $k$ as odd numbers and varied $k$ in $\left\{1, 3, 5, 7, 9 \right\}$. We left one language out and took the remaining languages as training examples to make predictions. This procedure was repeated for each language and the average prediction accuracies on all languages are reported.

\section{Detailed Many-to-Many Translation Results for LEE}\label{detailed_many_to_many_translation_results_for_lee}
\begin{figure}[t]
    \centering
    \includegraphics[width=0.4\textwidth]{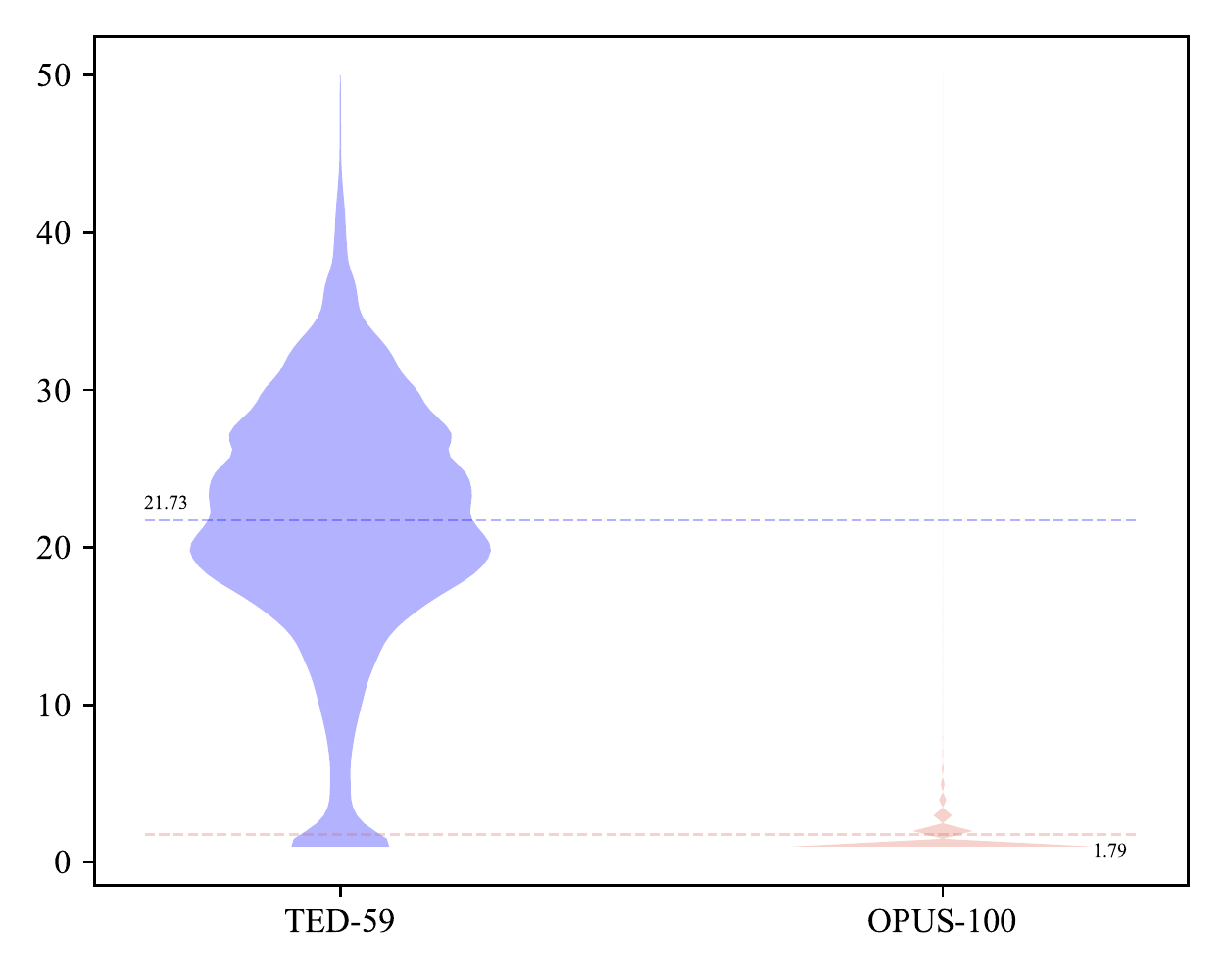}
    \caption{The distribution of the numbers of translations in other languages for each unique English sentence on the TED-59 and OPUS-100 training data. The dashed lines show the mean values.}\label{en_sentence_weight}
\end{figure}
Our findings summarized from Table~\ref{pre_mnmt_results} can be shown as follows:

\paragraph{The ways to indicate the desired target language to MNMT models can significantly affect the translation performance, especially on zero-shot translation directions.}
There are 0.33 and 5.88 maximum average BLEU differences on supervised (\circled{4} vs. \circled{6}) and zero-shot (\circled{6} vs. \circled{9}) translation directions respectively on the TED-59 dataset among the variations of LEE with the same number of parameters. Similarly, although LEE does not use extra parameters compared to Token$_{\mathrm{src}}$ and Token$_{\mathrm{tgt}}$, the average BLEU difference on zero-shot translation directions between LEE and Token$_{\mathrm{src}}$ can be significant on both datasets (6.56 on the TED-59, \circled{6} vs. \circled{2}) (3.31 on the OPUS-100 dataset, \circled{\small 11} vs. \circled{\small 12}). These indicate the importance of the effective target language information injecting method for MNMT models.

\paragraph{The performance gap between Token$_{\mathrm{src}}$ and Token$_{\mathrm{tgt}}$ on zero-shot translation directions varies substantially across different datasets.}
Token$_{\mathrm{src}}$ obtains the best performance in terms of both average BLEU and WR and outperforms Token$_{\mathrm{tgt}}$ by 8.08 BLEU on zero-shot translation directions on the TED-59 dataset (\circled{1} vs. \circled{2}). However, it lags behind Token$_{\mathrm{tgt}}$ on zero-shot translation directions on the OPUS-100 dataset (\circled{\small 10} vs. \circled{\small 11}). In order to find the main reasons behind the performance discrepancy between Token$_{\mathrm{src}}$ and Token$_{\mathrm{tgt}}$ on different datasets, we analyzed the two datasets from the perspective of their multi-way alignment since TED talks have been translated into many languages. Specifically, we counted the numbers of translations in other languages for each unique English sentence in the training data of the two datasets.\footnote{We collected English sentences by concatenating all parallel corpora of English and eliminating duplicate entries.} The results are visualized in Figure~\ref{en_sentence_weight}. Although both TED-59 and OPUS-100 are English-centric datasets, the majority of English sentences in the TED-59 dataset have more than one translations in other languages. We hypothesize that identical English sentences on the source side paired with distinct target sentences in different languages may encourage MNMT models to capture the correlations between the prepended tokens of Token$_{\mathrm{src}}$ and the target languages, which benefits zero-shot translation and is consistent with the finding of \citet{DBLP:conf/acl/WuCWL21}.
\paragraph{Prepending a target language token to the source side (Token$_{\mathrm{src}}$) and to the target side (Token$_{\mathrm{tgt}}$) benefit En $\rightarrow$ XX and XX $\rightarrow$ En translation directions respectively}
The average 0.5 BLEU gain and 91.38\% WR on the TED-59 dataset (\circled{1} vs. \circled{2}) together with the average 0.58 BLEU gain and 81.91\% WR on the OPUS-100 dataset (\circled{\small 10} vs. \circled{\small 11}) on the En $\rightarrow$ XX translation directions indicate that Token$_{\mathrm{src}}$ is more preferable for En $\rightarrow$ XX language pairs than Token$_{\mathrm{tgt}}$. Moreover, Token$_{\mathrm{src}}$ achieves the highest WR on En $\rightarrow$ XX translation directions across the two datasets. Despite the leading performance of Token$_{\mathrm{src}}$ on En $\rightarrow$ XX translation directions, it suffers from inferior average BLEU and WR on XX $\rightarrow$ En translation directions, which results in comparable average BLEU with Token$_{\mathrm{tgt}}$ over all supervised translation directions. In contrast, Token$_{\mathrm{tgt}}$ achieves better performance on XX $\rightarrow$ En translation directions than Token$_{\mathrm{src}}$.
\paragraph{Embodying the target language embedding in the decoder is preferable for supervised translation directions} The candidate positions for embodying language information are scattered across the encoder (position 1, 2) and decoder (position 3, 4, 5, 6). Table~\ref{pre_mnmt_results} shows that embodying the language embedding into positions in the decoder can achieve inprovements in average BLEU (\circled{1}\circled{2} vs. \circled{5}\circled{6}\circled{7}\circled{8}, \circled{\small 10}\circled{\small 11} vs. \circled{\small 12}), while incorporating the language embedding to positions in the encoder will cause performance drop (\circled{1}\circled{2} vs. \circled{3}\circled{4}). This suggests that embodying the language embedding for the target language into positions closer to target translations is the key ingredient to improve model performance on supervised translation directions for LEE.

\section{Linguistic Typology Prediction Results for LEE}\label{typology_results_for_lee}
\begin{table}[h]
\centering
\resizebox{0.46\textwidth}{!}{
    \begin{tabular}{c|c|c|l|ccccc|c}
    \toprule
        \multicolumn{1}{c}{\multirow{2}{*}{ID}} &
        \multicolumn{1}{c}{\multirow{2}{*}{Feature}} &\multicolumn{1}{c}{\multirow{2}{*}{Dataset}} & \multicolumn{1}{c}{\multirow{2}{*}{Model}} & \multicolumn{5}{c}{\multirow{1}{*}{k}} & \multicolumn{1}{c}{\multirow{2}{*}{Max}} \\
    \cmidrule{5-9}
        \multicolumn{1}{c}{} & \multicolumn{1}{c}{} & \multicolumn{1}{c}{} & \multicolumn{1}{c}{} & 1 & 3 & 5 & 7 & \multicolumn{1}{c}{9} & \multicolumn{1}{c}{} \\
    \midrule
        \circled{1} & \multicolumn{1}{c|}{\multirow{6}{*}{Syntax}} & TED-59 & Token$_{\mathrm{tgt}}$ & 81.56 & 82.15 & \textbf{82.81} & 81.18 & 80.79 & 82.81 \\
        \circled{2} &  & TED-59 & Token$_{\mathrm{src}}$ & 86.11 & \textbf{86.14} & 85.15 & 83.92 & 82.63 & \textbf{86.14} \\
        \circled{3} &  & TED-59 & LEE$_{4,5}$ & 82.40 & 84.70 & \textbf{85.59} & 84.45 & 83.42 & 85.59 \\
    \cmidrule{3-10}
        \circled{4} &  & OPUS-100 & Token$_{\mathrm{tgt}}$ & 82.66 & \textbf{83.97} & 83.49 & 83.25 & 82.98 & 83.97 \\
        \circled{5} &  & OPUS-100 & Token$_{\mathrm{src}}$ & \textbf{84.80} & 84.32 & 83.70 & 83.93 & 83.12 & \textbf{84.80} \\
        \circled{6} &  & OPUS-100 & LEE$_{4,5}$ & \textbf{84.42} & 83.43 & 83.80 & 82.34 & 82.07 & 84.42 \\
    \midrule
        \circled{7} & \multicolumn{1}{c|}{\multirow{6}{*}{Phonology}} & TED-59 & Token$_{\mathrm{tgt}}$ & 71.96 & \textbf{84.23} & 67.28 & 59.21 & 59.15 & 84.23 \\
        \circled{8} &  & TED-59 & Token$_{\mathrm{src}}$ & 80.11 & \textbf{84.97} & 67.28 & 59.03 & 59.03 & \textbf{84.97} \\
        \circled{9} &  & TED-59 & LEE$_{4,5}$ & 69.68 & \textbf{78.48} & 66.66 & 58.97 & 59.03 & 78.48 \\
    \cmidrule{3-10}
        \circled{\scriptsize 10} &  & OPUS-100 & Token$_{\mathrm{tgt}}$ & 79.57 & \textbf{85.94} & 81.38 & 73.78 & 73.82 & 85.94 \\
        \circled{\scriptsize 11} &  & OPUS-100 & Token$_{\mathrm{src}}$ & \textbf{86.76} & 86.50 & 83.94 & 73.95 & 73.95 & \textbf{86.76} \\
        \circled{\scriptsize 12} &  & OPUS-100 & LEE$_{4,5}$ & \textbf{85.79} & 80.74 & 73.82 & 73.87 & 74.03 & 85.79 \\
    \midrule
        \circled{\scriptsize 13} & \multicolumn{1}{c|}{\multirowcell{6}{Phonetic\\ inventory}} & TED-59 & Token$_{\mathrm{tgt}}$ & 86.67 & \textbf{87.86} & 87.11 & 85.07 & 65.80 & 87.86 \\
        \circled{\scriptsize 14} &  & TED-59 & Token$_{\mathrm{src}}$ & 87.85 & \textbf{88.38} & 87.44 & 85.32 & 66.06 & \textbf{88.38} \\
        \circled{\scriptsize 15} &  & TED-59 & LEE$_{4,5}$ & 86.52 & \textbf{88.03} & 87.15 & 85.12 & 65.73 & 88.03 \\
    \cmidrule{3-10}
        \circled{\scriptsize 16} &  & OPUS-100 & Token$_{\mathrm{tgt}}$ & \textbf{87.89} & 87.68 & 72.78 & 59.41 & 59.47 & 87.89 \\
        \circled{\scriptsize 17} &  & OPUS-100 & Token$_{\mathrm{src}}$ & \textbf{87.86} & 87.80 & 72.96 & 59.47 & 59.42 & 87.86 \\
        \circled{\scriptsize 18} &  & OPUS-100 & LEE$_{4,5}$ & \textbf{88.38} & 87.76 & 72.80 & 59.42 & 59.24 & \textbf{88.38} \\
    \bottomrule
    \end{tabular}
    }
\caption{Linguistic typology prediction accuracies on syntax, phonology and phonetic inventory features using the language embedding learned by Token$_{\mathrm{tgt}}$, Token$_{\mathrm{src}}$ and LEE$_{4,5}$ which are trained on the TED-59 and OPUS-100 datasets respectively. k denotes the number of nearest neighbors in k-NN. Max denotes the maximum accuracy when k varies in $\left\{1, 3, 5, 7, 9 \right\}$.}\label{lee_typology_results_appendix_table}
\end{table}

\section{Linguistic Typology Prediction Results for Models Trained on OPUS-100}\label{typology_results_opus_section}
\begin{figure*}[ht!]
    \centering
        \begin{subfigure}{0.325\textwidth}
            \centering
            \includegraphics[width=\linewidth]{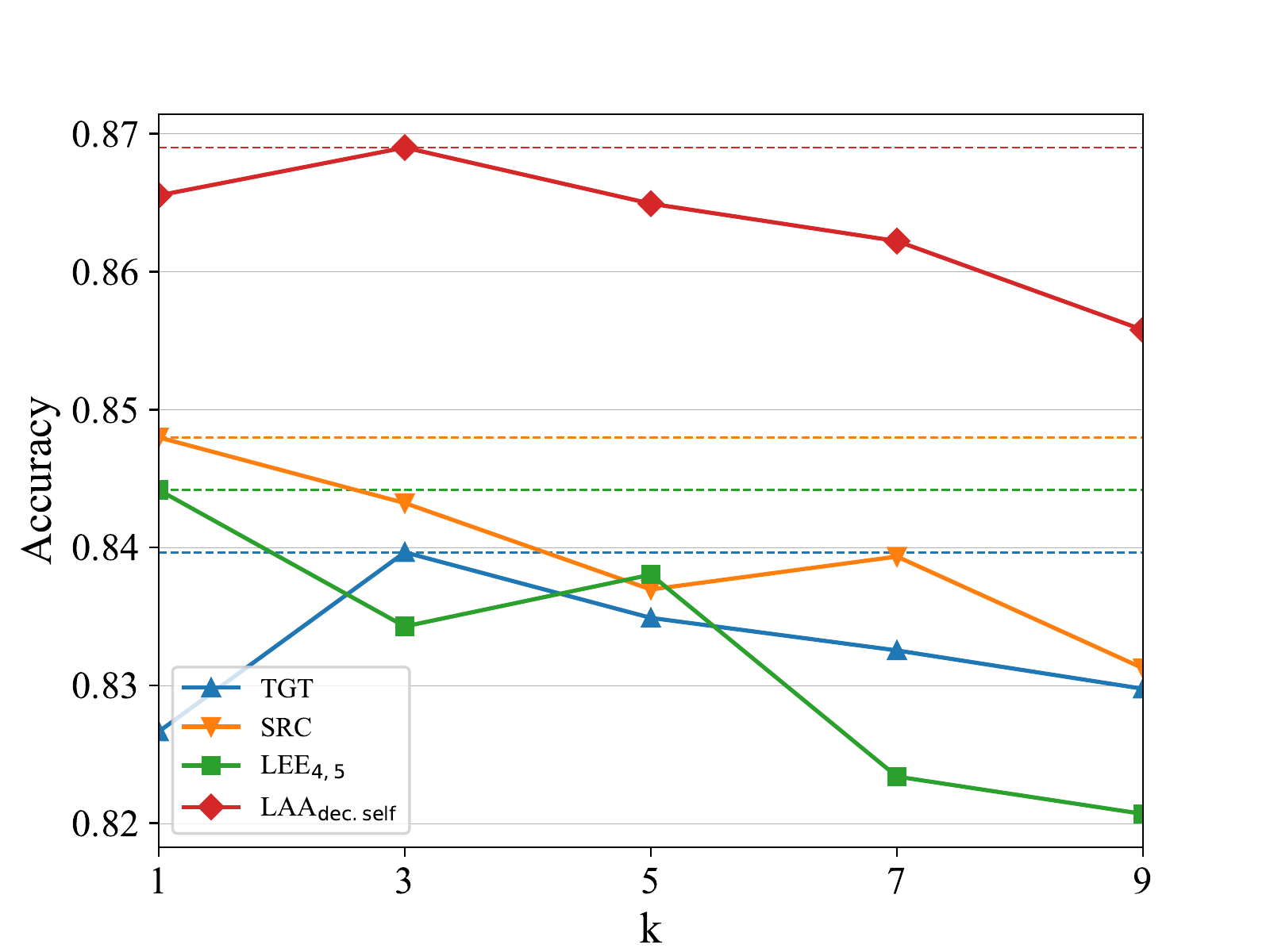}  
            \caption{Syntax}
        \end{subfigure}
        \begin{subfigure}{0.325\textwidth}
            \centering
            \includegraphics[width=\linewidth]{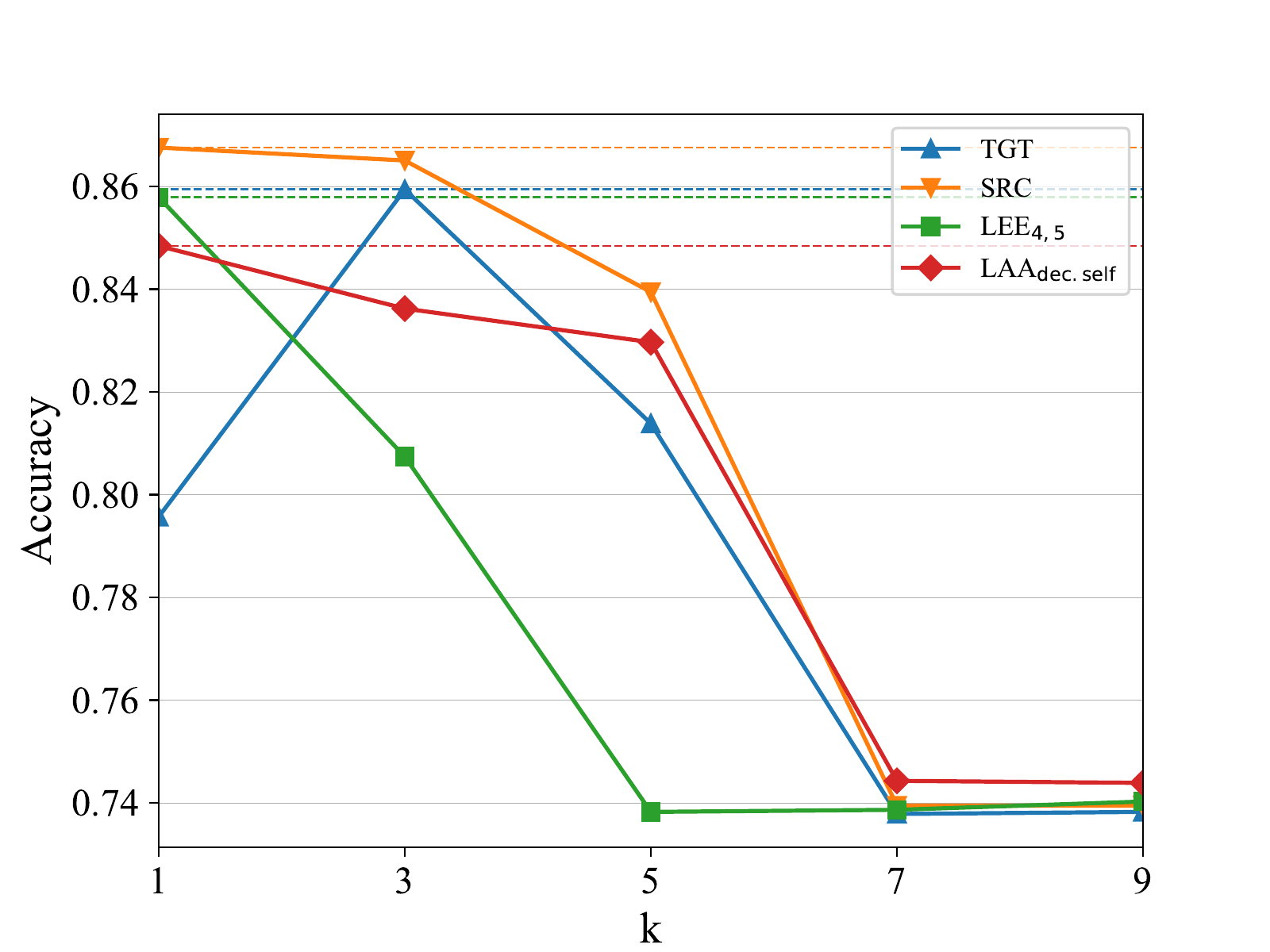}  
            \caption{Phonology}
        \end{subfigure}
        \begin{subfigure}{0.325\textwidth}
            \centering
            \includegraphics[width=\linewidth]{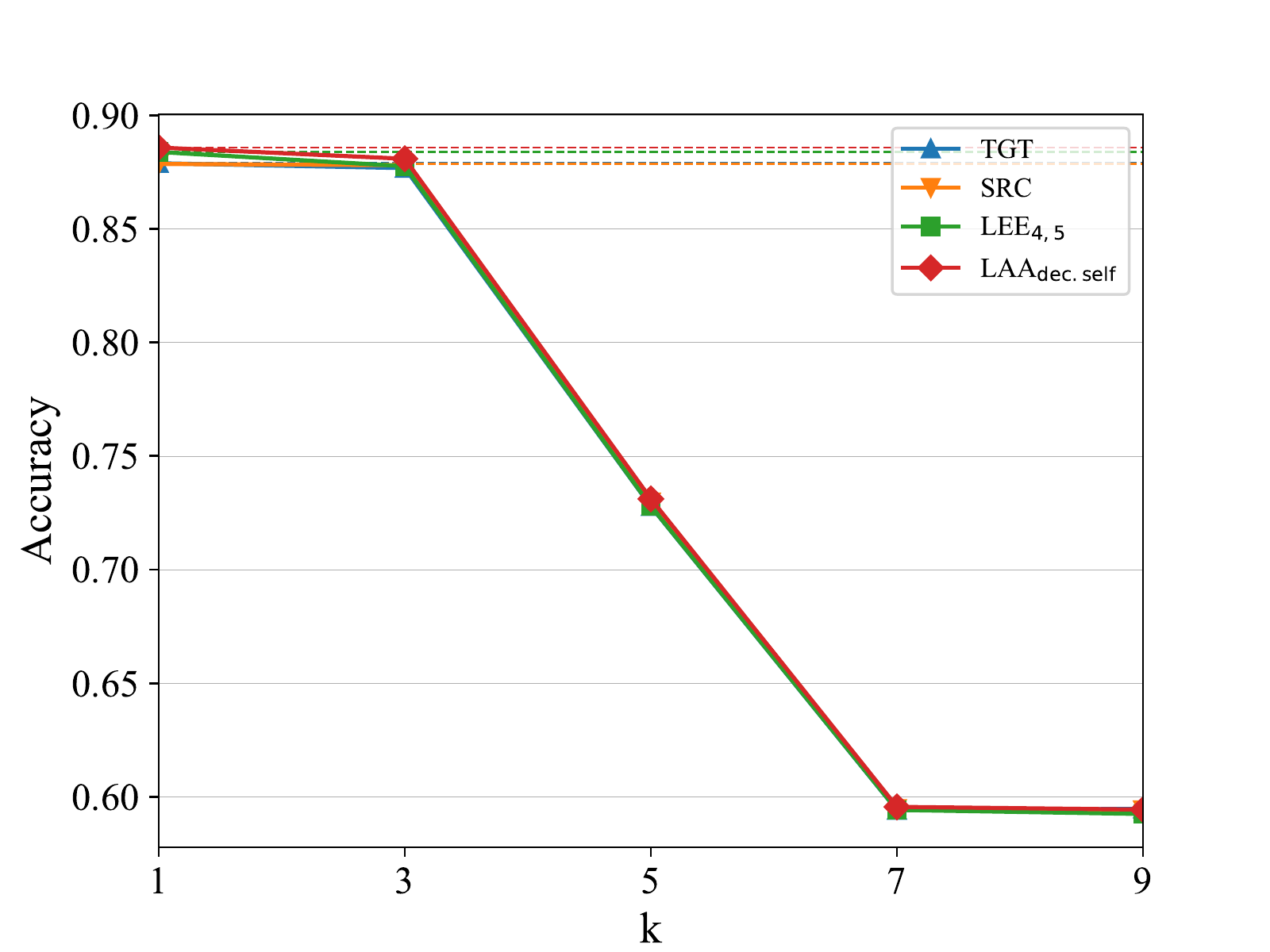}  
            \caption{Phonetic inventory}
        \end{subfigure}
    \caption{Prediction accuracy on syntax, phonology and phonetic inventory features using the language embeddings learned by Token$_{\mathrm{tgt}}$, Token$_{\mathrm{src}}$, LEE$_{4,5}$ and LAA$_{\mathrm{dec.self}}$ which are trained on the OPUS-100 dataset.}\label{typology_results_opus}
\end{figure*}

Figure \ref{typology_results_opus} shows the linguistic typology prediction results for language representations extracted from MNMT models which are trained on the OPUS-100 dataset.

\section{Many-to-Many Translation Results of LAA$_{\mathrm{dec.self}}^{R}$ on Raw Data Distribution}\label{mnmt_results_random_no_oversample_section}
Table \ref{mnmt_results_random_no_oversample} presents the many-to-many translation results of LAA$_{\mathrm{dec.self}}^{R}$ when oversampling is removed.

\begin{table*}[t]
\centering
    \resizebox{\textwidth}{!}{
        \begin{tabular}{c|c|l|c|ll|ll|ll|lll}
        \toprule
            \multicolumn{1}{c}{\multirow{2}{*}{ID}} & \multicolumn{1}{c}{\multirow{2}{*}{Dataset}} & \multicolumn{1}{c}{\multirow{2}{*}{Model}} & \multicolumn{1}{c}{\multirow{2}{*}{\#Param}} & \multicolumn{2}{c|}{En $\rightarrow$ XX} &  \multicolumn{2}{c|}{XX $\rightarrow$ En} & \multicolumn{2}{c|}{All} & \multicolumn{3}{c}{Zero-shot} \\
        \cmidrule{5-13}
            \multicolumn{1}{c}{} & \multicolumn{1}{c}{} & \multicolumn{1}{c}{} & \multicolumn{1}{c}{} & BLEU & WR & BLEU & WR & BLEU & WR & BLEU & LangAcc & WR \\
        \midrule
            \circled{1} & TED-59 & Token$_{\mathrm{tgt}}^{-}$ & 77M & 19.54 & \textit{ref} & 24.23 & \textit{ref} & 21.89 & \textit{ref} & 2.84 & 40.94 & \textit{ref} \\
            \circled{2} & TED-59 & Token$_{\mathrm{src}}^{-}$ & 77M & 20.25 & \textbf{96.55} & 23.65 & 8.62 & 21.95 & 52.59 & 9.65 & 65.45 & 96.77 \\
            \circled{3} & TED-59 & LAA$_{\mathrm{dec.self}}^{-}$ & 92M & 20.64 & \textbf{96.55} & 25.16 & \textbf{98.28} & \textbf{22.90} & \textbf{97.41} & 9.93 & 75.04 & 97.67 \\
            \circled{4} & TED-59 & LAA$_{\mathrm{dec.self}}^{-}$ + LEE$_{4,5}^{-}$ & 92M & \textbf{20.67} & 94.83 & 25.05 & 93.10 & 22.86 & 93.97 & \textbf{10.02} & \textbf{75.26} & \textbf{97.82} \\
            \circled{5} & TED-59 & LAA$_{\mathrm{dec.self}}^{R-}$ + Token$_{\mathrm{src}}^{-}$ & 92M & 19.65$\pm$0.01 & 61.38$\pm$2.61 & 25.75$\pm$0.01 & 90.34$\pm$1.54 & 22.70$\pm$0.01 & 75.86$\pm$0.86 & 9.89$\pm$0.00 & 73.30$\pm$0.02 & 95.91$\pm$0.06 \\
            \circled{6} & TED-59 & LAA$_{\mathrm{dec.self}}^{R-}$ + Token$_{\mathrm{tgt}}^{-}$ & 92M & 18.53$\pm$0.01 & 1.03$\pm$0.94 & \textbf{26.30$\pm$0.02} & 93.10$\pm$0.00 & 22.41$\pm$0.01 & 47.07$\pm$0.47 & 1.02$\pm$0.00 & 11.37$\pm$0.01 & 1.06$\pm$0.07 \\
        \midrule
            \circled{7} & OPUS-100 & Token$_{\mathrm{tgt}}^{-}$ & 77M & 21.82 & \textit{ref} & 28.45 & \textit{ref} & 25.14 & \textit{ref} & 6.63 & 58.76 & \textit{ref} \\
            \circled{8} & OPUS-100 & Token$_{\mathrm{src}}^{-}$ & 77M & 22.15 & 74.47 & 27.68 & 10.64 & 24.91 & 42.55 & 4.91 & 37.80 & 30.00 \\
            \circled{9} & OPUS-100 & LAA$_{\mathrm{dec.self}}^{-}$ & 103M & 23.57 & \textbf{91.49} & 28.71 & 70.21 & 26.14 & 80.85 & 11.93 & 80.39 & \textbf{100.00} \\
            \circled{\small 10} & OPUS-100 & LAA$_{\mathrm{dec.self}}^{-}$ + LEE$_{4,5}^{-}$ & 103M & \textbf{23.69} & \textbf{91.49} & \textbf{28.88} & \textbf{78.72} & \textbf{26.29} & \textbf{85.11} & \textbf{12.77} & \textbf{85.00} & \textbf{100.00} \\
            \circled{\small 11} & OPUS-100 & LAA$_{\mathrm{dec.self}}^{R-}$ + Token$_{\mathrm{src}}^{-}$ & 103M & 21.63$\pm$0.01 & 50.21$\pm$1.39 & 27.07$\pm$0.02 & 0.64$\pm$0.58 & 24.35$\pm$0.01 & 25.43$\pm$0.69 & 4.96$\pm$0.02 & 41.21$\pm$0.05 & 31.33$\pm$1.83 \\
            \circled{\small 12} & OPUS-100 & LAA$_{\mathrm{dec.self}}^{R-}$ + Token$_{\mathrm{tgt}}^{-}$ & 103M & 20.75$\pm$0.02 & 9.15$\pm$1.61 & 28.11$\pm$0.03 & 20.00$\pm$1.75 & 24.43$\pm$0.02 & 14.57$\pm$0.97 & 6.80$\pm$0.02 & 66.74$\pm$0.07 & 52.00$\pm$1.83 \\
        \bottomrule
        \end{tabular}
    }
\caption{Experiment results of LAA$_{\mathrm{dec.self}}^{R-}$ on the two datasets. The superscript "-" denotes that there is no oversampling. The results of \circled{1}\circled{2}\circled{3}\circled{4}\circled{7}\circled{8}\circled{9}\circled{\small 10} are from Table~\ref{mnmt_results_no_over_sample}.}\label{mnmt_results_random_no_oversample}
\end{table*}

\end{document}